\documentclass[10pt,twocolumn,letterpaper]{article}

\usepackage[pagenumbers]{wacv} %

\usepackage{tikz}
\usetikzlibrary{trees,arrows.meta,positioning}
\usepackage[ruled,vlined]{algorithm2e}
\usepackage{amsmath}
\usepackage{comment}

\usepackage{graphicx}
\usepackage{amsmath}
\usepackage{amssymb}
\usepackage{booktabs}
\usepackage{multirow}

\definecolor{cvprblue}{rgb}{0.21,0.49,0.74}
\usepackage[pagebackref,breaklinks,colorlinks,allcolors=cvprblue]{hyperref}

\usepackage[capitalize]{cleveref}
\crefname{section}{Sec.}{Secs.}
\Crefname{section}{Section}{Sections}
\Crefname{table}{Table}{Tables}
\crefname{table}{Tab.}{Tabs.}
\usepackage[accsupp]{axessibility}  %
\def\wacvPaperID{260} %
\def\confName{WACV}
\def\confYear{2027}

\title{From Geometry to Function$\colon$ A Multi-Task LiDAR Benchmark Dataset for Semantic and Functional Parsing of Urban Facades}

\author{
Abdiaziz Muse$^{1}$ \quad Tooba Javed$^{1}$ \quad Tao Pu$^{2}$ \quad
Jiju Poovvancheri$^{1}$ \quad Dong Chen$^{2}$\\[2pt]
$^{1}$Graphics and Spatial Computing Lab, Saint Mary's University, Halifax, Canada\\
$^{2}$Nanjing Forestry University, Nanjing, China\\
{\tt\small \{abdiaziz.muse, tooba.javed, jiju.poovvancheri\}@smu.ca \quad
\{taopu980215, chendong\}@njfu.edu.cn}
}

\begin{document}
\maketitle
\begin{abstract}
Understanding building facades is central to urban digital twins, heritage modelling, and sustainable city analysis, yet existing LiDAR benchmarks stop at geometry and semantic segmentation and offer little support for reasoning about the functional role of each architectural element. We introduce HFX3D, a large-scale urban facade benchmark of approximately 492 million LiDAR points from 26 buildings in Halifax, Canada, captured across four seasons, day and night, and varied weather. Each facade is annotated at three tiers, semantic, instance, and per-instance functional attributes, making it, to our knowledge, the first facade dataset to pair geometry and semantics with functional labels. Annotations are produced by a semi-automatic pipeline that combines geometry-guided heuristics with expert validation. We benchmark representative point-based, graph-based, convolutional, and transformer architectures and find that semantic segmentation is already challenging under severe class imbalance, while instance separation of repetitive structures and functional reasoning from geometry are substantially harder. The benchmark establishes a new foundation for function-aware 3D urban understanding. Project page: \url{https://jijup.github.io/hfx3d-benchmark/}; code: \url{https://github.com/jijup/hfx3d-benchmark}.
\end{abstract}
    
\section{Introduction}

\label{sec:intro}

\begin{figure}[h]
\centering
\includegraphics[width=\linewidth]{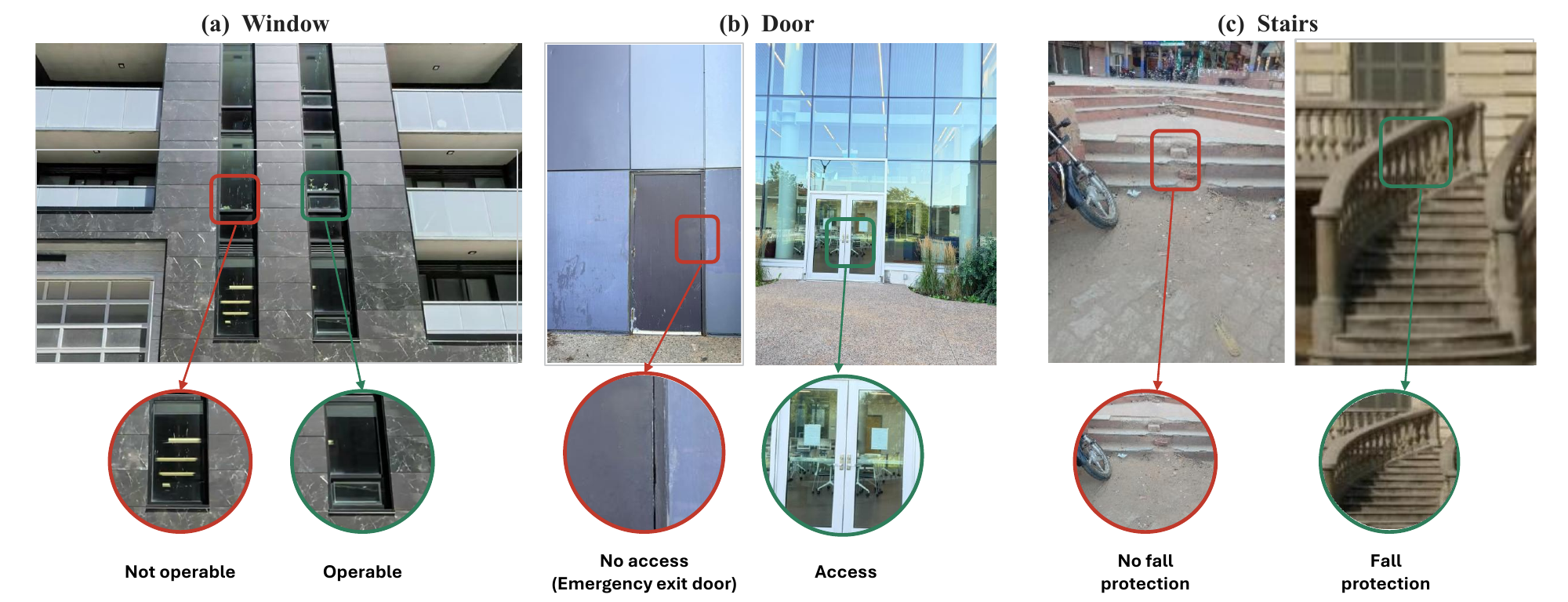}
\caption{\textbf{Functional variation within semantic classes.}
Instances sharing the same semantic label can exhibit contrasting functional attributes: (a) window operability, (b) door access, and (c) stair fall protection. Red/green denote absent/present attributes; circles show zoomed regions. All three cases are present in HFX3D; the stair photographs in (c) are reproduced from~\cite{tylecek2013spatial,datacluster_stairs} for clearer illustration.}
\label{fig:attr_examples}
\end{figure}

Accurate 3D facade modelling is central to urban digital twins and building information modelling (BIM)~\cite{biljecki2015applications, kolbe2005citygml}. Recent advances in point cloud segmentation~\cite{qi2017pointnet, qi2017pointnetpp, wang2019dynamic, zhao2021point} can reliably label common facade components such as walls, windows, and doors, but semantic labels alone are insufficient for high-level applications. The functional role of each element, whether a window is operable, whether a balcony provides access, whether a column is load-bearing, is what drives structural analysis, energy modelling, and BIM reconstruction~\cite{musialski2013survey, martinovic2015bayesian}. Figure~\ref{fig:attr_examples} illustrates this within-class ambiguity: facade elements with the same semantic label can exhibit contrasting functional roles. Mobile LiDAR scanning now yields dense, RGB-coloured point clouds that capture facade geometry at centimetre resolution~\cite{tan2020toronto3d, chen2026cityfacade}, yet no existing 3D facade dataset provides functional annotations alongside semantic and instance labels.

Recent facade point cloud benchmarks have advanced the field along complementary directions. ZAHA~\cite{wysocki2024zahaintroducinglevelfacade} introduced the hierarchical Level of Facade Generalization (LoFG) for fine-grained semantic parsing, City-Facade~\cite{chen2026cityfacade} extended facade understanding to city-scale mobile LiDAR with semantic and instance annotations, and UnderOneFacade~\cite{wang2026underonefacade} emphasized geographic diversity and cross-domain generalization. These benchmarks improve semantic granularity, instance understanding, and transferability, but do not explicitly capture the functional roles of individual facade elements, leaving a key gap for downstream BIM and digital-twin reasoning.
\begin{figure*}[t]
\centering
\includegraphics[width=\linewidth]{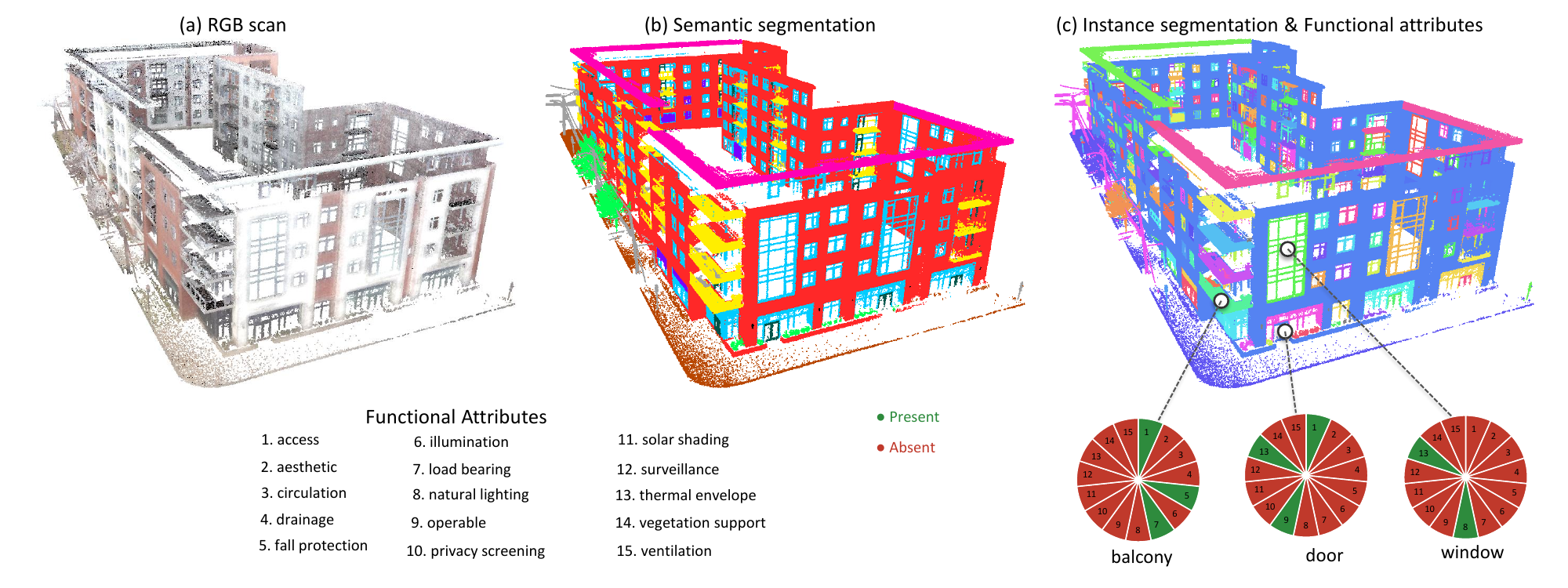}
\caption{\textbf{Three tiers of facade annotation.}
From a raw MLS scan we provide (a) registered RGB, (b) 12-class semantic
segmentation, and (c) per-instance segmentation. Each instance additionally
carries a 15-attribute functional vector, shown as radial glyphs for a
representative balcony, door, and window (green = present, red = absent).}
\label{fig:teaser}
\end{figure*}

We introduce HFX3D, a large-scale LiDAR facade dataset with three tiers of annotation (Figure~\ref{fig:teaser}): per-point semantic labels across 12 classes, instance-level grouping of individual facade elements, and 15 functional attributes per instance such as load-bearing and operable. It comprises approximately 492 million points from 26 buildings across five neighbourhoods of Halifax, Canada. These buildings were captured across four seasons, day and night, and varied weather, an environmental range rarely found in existing facade benchmarks. Functional attributes are grounded in architectural standards~\cite{ISO16739_2024,petersen1990art}. They are assigned through a semi-automatic pipeline that combines ontology-driven suggestions, geometry-guided refinement, and human validation against the corresponding RGB scans and publicly available street-level imagery. Our key contributions are listed below.

\noindent\textbf{Contributions.}
\begin{itemize}
    \setlength{\itemsep}{0.2em}
    \setlength{\parskip}{0pt}
    \setlength{\parsep}{0pt}
    \setlength{\topsep}{0.3em}
    \item To our knowledge, the first and largest RGB-coloured LiDAR facade benchmark to provide semantic, instance, and per-instance functional annotations.
    \item A scalable semi-automatic annotation pipeline that pairs geometry-guided instance generation with IFC- and Getty~AAT-grounded functional labeling, refined by geometric and contextual evidence and human validation.
    \item A unified, full-coverage benchmark of seven semantic backbones and seven instance methods spanning learned grouping and classical clustering, together with a cross-city generalization study against the ZAHA facade benchmark~\cite{wysocki2024zahaintroducinglevelfacade}.
\end{itemize}

\section{Related Work}

\newcommand{\tabyes}{\textcolor[RGB]{34,139,34}{\checkmark}}
\newcommand{\tabno}{\textcolor[RGB]{185,28,28}{$\times$}}
\newcommand{\tabpart}{\textcolor[RGB]{200,130,0}{($\checkmark$)}}
\begin{table*}[!t]
\centering
\caption{\textbf{Comparison of representative 3D point cloud segmentation datasets.}
Pts~=~total points (M); Avg~=~mean points per building/facade (M);
Cls~=~semantic classes; Nbhd~=~distinct neighbourhoods or districts captured;
Sem.~=~semantic; Inst.~=~instance; Func.~=~functional labels;
Inten.~=~LiDAR intensity per point included in released data;
Seas~=~seasons captured; Wthr~=~weather variation; D/N~=~day \& night captures;
Rev.~=~independent annotation review with inter-annotator agreement recorded.
`---'~=~not reported in original publication;
($\checkmark$)~=~available for part of the dataset only.
$^\dagger$Synthetic (mesh-based).}
\label{tab:facade_datasets}
\resizebox{\textwidth}{!}{%
\begin{tabular}{lc l c rr ccccc ccc c cccc}
\toprule
\textbf{Dataset} & \textbf{Year} & \textbf{Location} & \textbf{Nbhd} &
\textbf{Pts (M)} & \textbf{Avg (M)} & \textbf{Scenes} &
\textbf{Cls} & \textbf{Sem.} & \textbf{Inst.} & \textbf{Func.} &
\textbf{Sensor} & \textbf{RGB} & \textbf{Inten.} & \textbf{Facade} &
\textbf{Seas} & \textbf{Wthr} & \textbf{D/N} & \textbf{Rev.} \\
\midrule
\multicolumn{19}{l}{\textit{General outdoor / indoor benchmarks}} \\
\midrule
S3DIS~\cite{armeni2016cvpr}                  & 2016 & Stanford, USA                    & 1   & 696   & ---  & 272 rooms & 13  & \tabyes & \tabyes & \tabno & RGB-D & \tabyes & \tabno  & \tabno & --- & --- & --- & --- \\
ScanNet~\cite{dai2017scannet}                & 2017 & Various (indoor)                 & --- & ---   & ---  & 1,513     & 20  & \tabyes & \tabyes & \tabno & RGB-D & \tabyes & \tabno  & \tabno & --- & --- & --- & --- \\
Semantic3D~\cite{hackel2017isprs}            & 2017 & Switzerland / Austria            & --- & 4,000 & ---  & $\sim$30  &  8  & \tabyes & \tabno  & \tabno & TLS   & \tabyes & \tabyes & \tabno &  1 & --- & --- & --- \\
Paris-Lille-3D~\cite{DBLP:journals/corr/abs-1712-00032} & 2018 & Paris + Lille, France & 2 & 143 & ---   & 2\,km    & 50  & \tabyes & \tabno  & \tabno & MLS   & \tabno  & \tabyes & \tabno &  1 & --- & --- & --- \\
SemanticKITTI~\cite{behley2019semantickitti} & 2019 & Karlsruhe, Germany               & 1   & ---   & ---  & 22 seqs.  & 28  & \tabyes & \tabno  & \tabno & MLS   & \tabno  & \tabyes & \tabno &  1 & --- & --- & --- \\
Toronto-3D~\cite{tan2020toronto3d}           & 2020 & Toronto, Canada                  & 1   & 78    & ---  & 1\,km     &  8  & \tabyes & \tabno  & \tabno & MLS   & \tabyes & \tabyes & \tabno &  1 & --- & --- & --- \\
SensatUrban~\cite{hu2021sensaturban}         & 2021 & Birmingham, Cambridge, York, UK  & 3   & 3,000 & ---  & 3 cities  & 13  & \tabyes & \tabno  & \tabno & UAV   & \tabyes & \tabno  & \tabno &  1 & --- & --- & --- \\
KITTI-360~\cite{Liao2022PAMI}               & 2022 & Karlsruhe, Germany               & 1   & ---   & ---  & 73\,km    & 19  & \tabyes & \tabyes & \tabno & MLS   & \tabyes & \tabyes & \tabno &  1 & --- & --- & --- \\
UrbanBIS~\cite{yang2023urbannis}             & 2023 & Shenzhen + Qingdao + Wuhu, China & 3   & 2,500 & ---  & 6 scenes  & --- & \tabyes & \tabyes & \tabno & UAV   & \tabyes & ---     & \tabno & --- & --- & --- & --- \\
\midrule
\multicolumn{19}{l}{\textit{Facade-specific benchmarks}} \\
\midrule
ArCH~\cite{matrone2020arch}                                         & 2020 & Italy + France (heritage) & --- & 135 &  7.9 & 17 & 10 & \tabyes & \tabno & \tabno & TLS    & \tabyes & ---     & \tabyes & --- & --- & --- & --- \\
TUM-FACADE~\cite{wysockiTUMFACADE}                             & 2022 & Munich, Germany           &  1  & 118 &  8.4 & 14 & 17 & \tabyes & \tabno & \tabno & MLS    & \tabno  & \tabpart & \tabyes &  1 & --- & --- & --- \\
ARCH2S$^\dagger$~\cite{cheung2024arch2sdatasetbenchmarkchallenges} & 2024 & Hong Kong (synthetic)     &  1  & --- &  5.0 &  5 & 14 & \tabyes & \tabno & \tabno & Synth. & \tabyes & \tabno  & \tabyes & --- & --- & --- & --- \\
ZAHA~\cite{wysocki2024zahaintroducinglevelfacade}                   & 2024 & Munich, Germany           &  1  & 601 &  9.1 & 66 & 15 & \tabyes & \tabno & \tabno & MLS    & \tabno  & \tabno  & \tabyes &  1 & --- & --- & --- \\
City-Facade~\cite{chen2026cityfacade}                              & 2026 & Xiamen, China             &  1  & 200 &  --- & ---&  9 & \tabyes & \tabyes & \tabno & MLS    & \tabyes & \tabyes & \tabyes & --- & --- & --- & --- \\
UnderOneFacade~\cite{wang2026underonefacade}                       & 2026 & UK + Germany + Singapore &  3  & 2700 &  7.4 & 366 & 15 & \tabyes & \tabno  & \tabno & TLS+MLS & \tabpart & \tabpart & \tabyes & --- & --- & --- & --- \\
\midrule
\textbf{HFX3D (Ours)} & \textbf{2026} & Halifax, Canada & \textbf{5} &
  \textbf{492} & \textbf{18.9} & \textbf{26} & \textbf{12} &
  \tabyes & \tabyes & \tabyes & \textbf{MLS} & \tabyes & \tabyes & \tabyes &
  \textbf{4} & \tabyes & \tabyes & \tabyes \\
\bottomrule
\end{tabular}%
}
\end{table*}

\subsection{Facade Segmentation Datasets}
Facade parsing has moved from 2D image benchmarks such as eTRIMS~\cite{korc2009etrims} and Graz50~\cite{riemenschneider2012irregular} to dense 3D point clouds. General-purpose outdoor benchmarks such as Semantic3D~\cite{hackel2017isprs}, Paris-Lille-3D~\cite{DBLP:journals/corr/abs-1712-00032}, and KITTI-360~\cite{Liao2022PAMI} provide large annotated point clouds but treat buildings as coarse semantic units, while indoor benchmarks such as S3DIS~\cite{armeni2016cvpr} and ScanNet~\cite{dai2017scannet} advanced instance segmentation in controlled settings.

Dedicated facade datasets have emerged more recently. ArCH~\cite{matrone2020arch} is a pioneering heritage benchmark with detailed architectural classes, and TUM-FACADE~\cite{wysockiTUMFACADE}, extending TUM-MLS-2016~\cite{zhu2020tum}, advances campus-scale fine-grained labeling. ARCH2S~\cite{cheung2024arch2sdatasetbenchmarkchallenges} contributes controlled synthetic scenes well suited to pretraining. ZAHA~\cite{wysocki2024zahaintroducinglevelfacade} is the most comprehensive geometric facade benchmark to date, and City-Facade~\cite{chen2026cityfacade} takes a further step by adding instance-level annotation. The concurrent UnderOneFacade~\cite{wang2026underonefacade} assembles a cross-continental benchmark spanning the UK, Germany, and Singapore with a hierarchical semantic taxonomy, but does not provide instance or functional labels. Building on these efforts, HFX3D introduces per-instance functional attributes, as summarised in Table~\ref{tab:facade_datasets}.

\subsection{Deep Learning for 3D Facade Segmentation}
Point-based networks progressed from PointNet~\cite{qi2017pointnet} and PointNet++~\cite{qi2017pointnetpp} to graph-convolutional DGCNN~\cite{wang2019dynamic}. Later work added kernel-point KPConv~\cite{thomas2019kpconv}, sampling-based RandLA-Net~\cite{hu2020randla} for large outdoor scenes, and self-attention Point Transformers~\cite{wu2022ptv2,wu2024ptv3}. Despite this progress, facade segmentation remains challenging: severe class imbalance and the geometric similarity between planar elements leave rare classes with low recall.
\enlargethispage{2\baselineskip}

\subsection{Instance and Functional Facade Understanding}
Beyond per-point labels, urban digital twins and BIM require grouping points into element instances and inferring their functional role. Instance segmentation is well established indoors, through methods such as PointGroup~\cite{jiang2020pointgroup}, SoftGroup~\cite{vu2022softgroup}, and ISBNet~\cite{ngo2023isbnet} on ScanNet, while outdoor panoptic benchmarks such as SemanticKITTI~\cite{behley2019semantickitti} target dynamic objects; instance-level facade parsing, separating repetitive windows or adjacent balconies, remains comparatively underexplored.

Functional roles are encoded in standards such as CityGML, IFC, and Getty AAT~\cite{groger2012citygml, ISO16739_2024, petersen1990art}, which attach properties such as load-bearing role, thermal-envelope membership, and accessibility to building elements. However, to the best of our knowledge, existing LiDAR facade benchmarks do not provide per-instance functional attribute annotations. Bridging this gap, HFX3D brings these functional concepts into a unified large-scale LiDAR benchmark with semantic, instance, and per-instance functional annotations.

\section{Dataset Acquisition and Annotation}
\label{sec:dataset}

\subsection{Point Cloud Acquisition}

We captured 26 buildings with a GeoSLAM ZEB~Horizon handheld mobile LiDAR scanner~\cite{zebhorizon} and a co-mounted ZEB~Vision camera that supplies per-point RGB colour. The buildings were drawn from 50 surveyed across five neighbourhoods of Halifax, Nova Scotia, Canada, with the final subset chosen to capture diversity in facade complexity, building height, construction material, and architectural era. An operator walked each facade at ground level, following a closed-loop trajectory where access allowed. Capture spanned six to eight months and all four seasons, day and night, under snow, fog, and clear sun, an environmental range not reported by any existing facade benchmark in Table~\ref{tab:facade_datasets}.

Each cloud was reconstructed and denoised in GeoSLAM Faro~Connect~\cite{faroconnect}, then cleaned, coloured, and exported in CloudCompare~\cite{girardeau2011cloudcompare}.

\subsection{Semantic Annotation}

\begin{figure}[t]
\centering
\includegraphics[width=\linewidth]{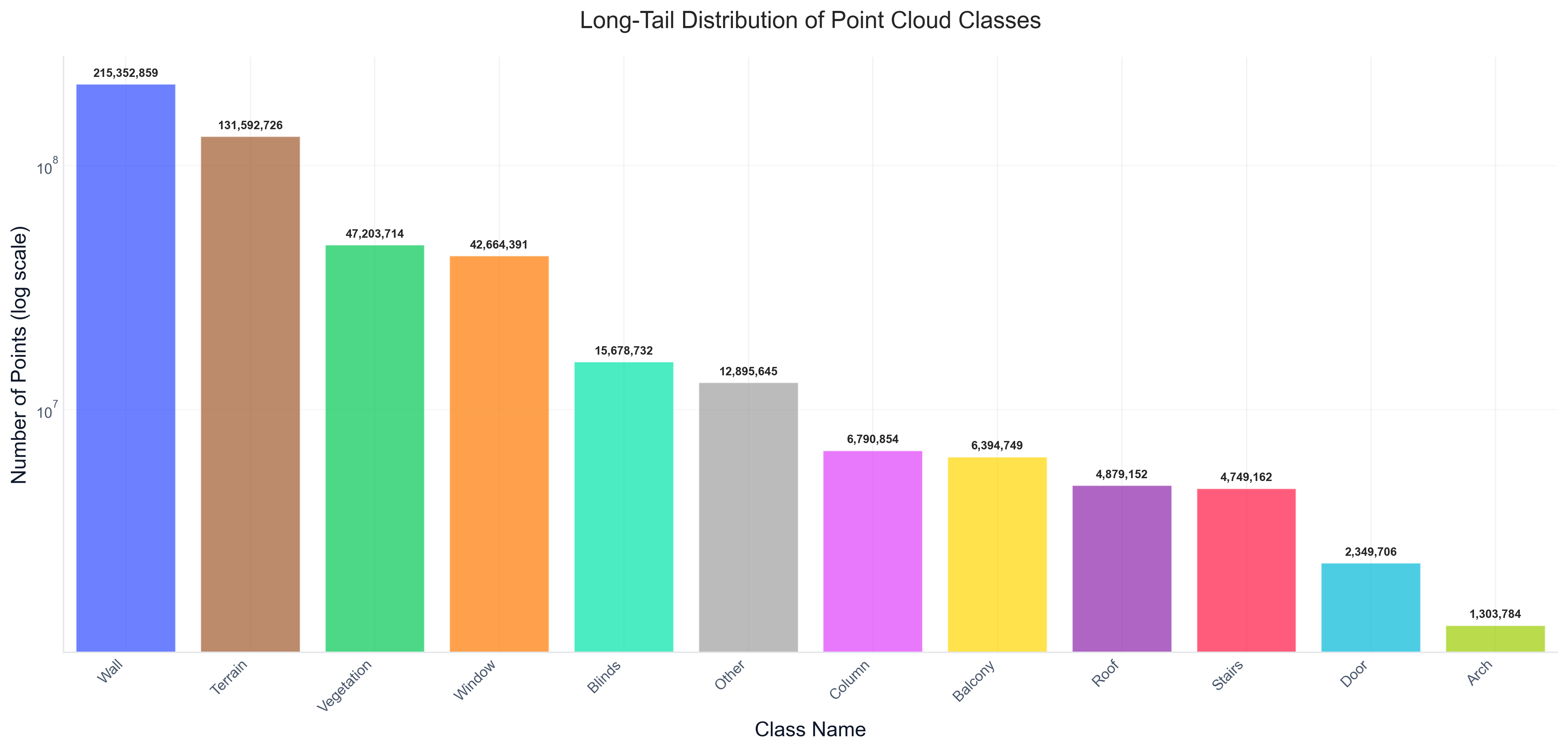}
\caption{\textbf{Long-tailed class distribution (log scale).} Wall and terrain dominate, while fine elements such as door and arch are far rarer.}
\label{fig:class_dist}
\vspace{-10pt}
\end{figure}

Four trained annotators manually annotated the point clouds in CloudCompare, using the co-registered ZEB~Vision fisheye imagery as visual reference. Each building was assigned to a primary annotator. We define 12 semantic classes: 0:~wall, 1:~window, 2:~door, 3:~balcony, 4:~vegetation, 5:~stairs, 6:~terrain, 7:~roof, 8:~blinds, 9:~other, 10:~column, and 11:~arch. As shown in Figure~\ref{fig:class_dist}, the resulting point distribution is highly imbalanced, with wall and terrain accounting for the majority of labeled points. Annotation was particularly challenging in regions containing geometrically or visually ambiguous elements, partial occlusions from vegetation and adjacent structures, and sparse LiDAR returns from reflective surfaces such as glass, a known difficulty for facade segmentation~\cite{wysocki2024zahaintroducinglevelfacade}. These ambiguities were further exacerbated under adverse acquisition conditions, including snow cover and low illumination. To promote consistency across buildings and annotators, all annotations followed predefined class definitions and were periodically reviewed for adherence to the annotation protocol.

Every building then underwent two rounds of cross-review by annotators who had not originally labeled it, with disagreements resolved by majority vote. For each building we logged the IoU against the original labels, Cohen's Kappa~\cite{cohen1960coefficient}, and the number of regions added, deleted, or modified, applying first-round corrections before the second. After first-round corrections were applied, second-round mean Cohen's Kappa reached 0.99; the most frequent residual correction in the second round was confusion between wall and window labels, and window remains the hardest of the frequent classes for every semantic segmentation baseline in Section~\ref{sec:benchmark}.

\subsection{Instance Segmentation via Clustering}
\label{subsec:instance_clustering}

To obtain instances, we cluster each semantic class with DBSCAN~\cite{ester1996density}, with $\epsilon$ tuned per building from 0.09 to 0.7 so that individual elements, such as a single window in a glass pane or a single bush, are separated into distinct instances. This automatic step produced 8{,}586 candidate instances across the 26 buildings; annotators then refined these segments in CloudCompare, merging over-segmented and splitting under-segmented elements and correcting semantic labels in the same pass, yielding 6{,}418 final instances, a 25.3\% reduction that reflects DBSCAN's tendency to over-segment rather than under-segment on this data. One building required net splitting rather than merging, confirming that review corrected in both directions rather than uniformly pruning clusters. Residual unassigned points were eliminated after refinement for all but two buildings. Each point then carries an instance ID alongside its semantic label.

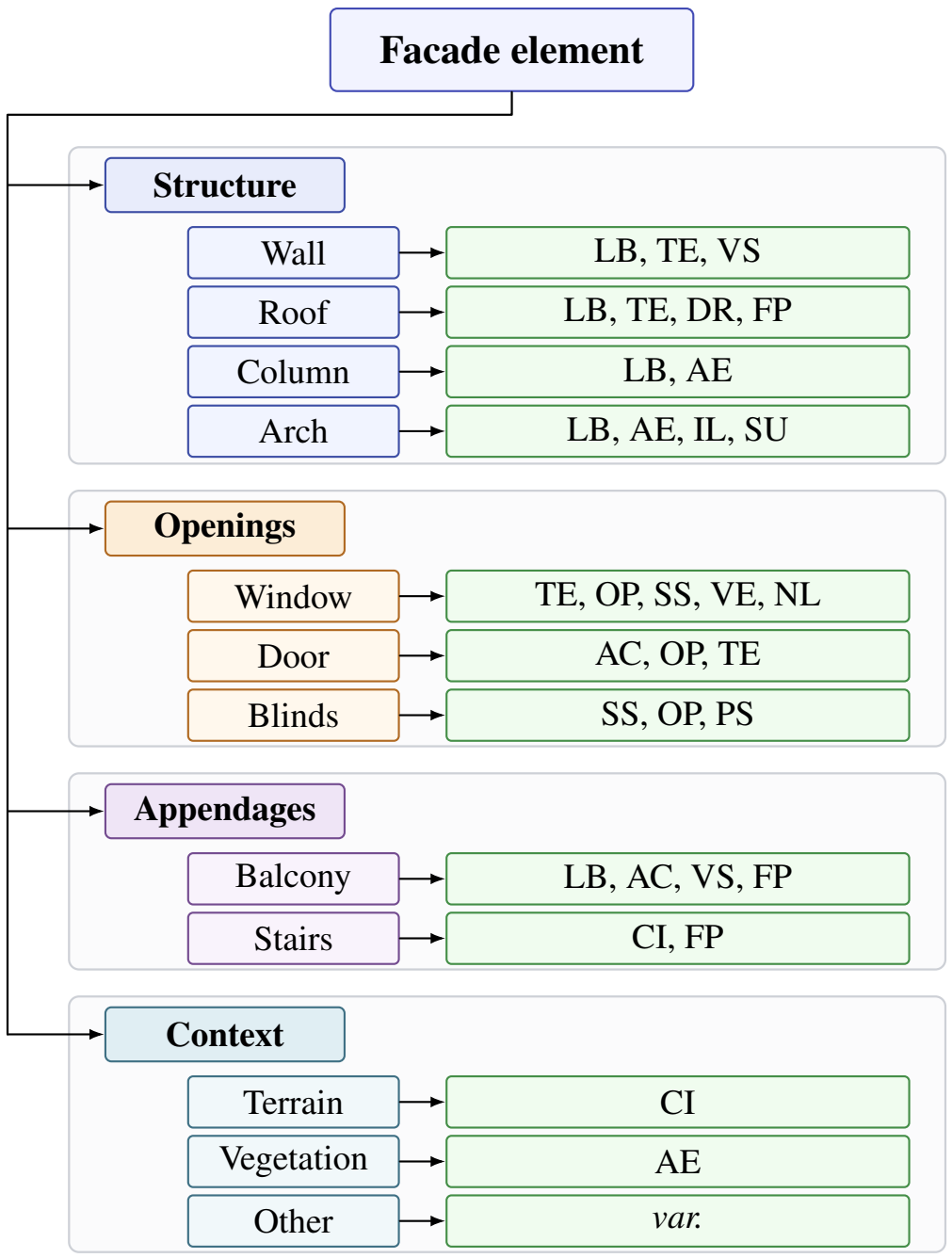
\begin{figure}[t]
\centering

\definecolor{RootBorder}{RGB}{66,75,180}
\definecolor{RootFill}{RGB}{242,243,255}

\definecolor{PanelBorder}{RGB}{205,208,214}
\definecolor{PanelFill}{RGB}{251,251,252}

\definecolor{StructureBorder}{RGB}{61,78,165}
\definecolor{StructureFill}{RGB}{241,244,255}
\definecolor{StructureHead}{RGB}{232,236,251}

\definecolor{OpeningsBorder}{RGB}{176,106,35}
\definecolor{OpeningsFill}{RGB}{255,248,237}
\definecolor{OpeningsHead}{RGB}{252,237,215}

\definecolor{AppendagesBorder}{RGB}{116,76,144}
\definecolor{AppendagesFill}{RGB}{249,243,252}
\definecolor{AppendagesHead}{RGB}{239,229,246}

\definecolor{ContextBorder}{RGB}{63,112,132}
\definecolor{ContextFill}{RGB}{241,248,250}
\definecolor{ContextHead}{RGB}{224,238,243}

\definecolor{FunctionBorder}{RGB}{68,142,71}
\definecolor{FunctionFill}{RGB}{240,252,239}

\begin{tikzpicture}[
x=1.08cm,
y=0.95cm,
font=\scriptsize,
spine/.style={
    draw=black,
    line width=0.42pt,
    line cap=round,
    line join=round
},
flow/.style={
    draw=black,
    -{Latex[length=1.05mm,width=0.78mm]},
    line width=0.40pt,
    line cap=round
},
rootbox/.style={
    draw=RootBorder,
    fill=RootFill,
    rounded corners=1.4pt,
    minimum width=2.55cm,
    minimum height=0.58cm,
    inner xsep=3pt,
    inner ysep=1.8pt,
    align=center,
    font=\footnotesize\bfseries
},
panel/.style={
    draw=PanelBorder,
    fill=PanelFill,
    rounded corners=2pt,
    line width=0.42pt
},
groupbase/.style={
    rounded corners=1.2pt,
    minimum width=1.68cm,
    minimum height=0.38cm,
    inner xsep=2pt,
    inner ysep=1.3pt,
    align=center,
    font=\scriptsize\bfseries
},
structuregroup/.style={
    groupbase,
    draw=StructureBorder,
    fill=StructureHead
},
openingsgroup/.style={
    groupbase,
    draw=OpeningsBorder,
    fill=OpeningsHead
},
appendagesgroup/.style={
    groupbase,
    draw=AppendagesBorder,
    fill=AppendagesHead
},
contextgroup/.style={
    groupbase,
    draw=ContextBorder,
    fill=ContextHead
},
classbase/.style={
    rounded corners=1pt,
    minimum width=1.48cm,
    minimum height=0.36cm,
    inner xsep=2pt,
    inner ysep=1.2pt,
    align=center
},
structureclass/.style={
    classbase,
    draw=StructureBorder,
    fill=StructureFill
},
openingsclass/.style={
    classbase,
    draw=OpeningsBorder,
    fill=OpeningsFill
},
appendagesclass/.style={
    classbase,
    draw=AppendagesBorder,
    fill=AppendagesFill
},
contextclass/.style={
    classbase,
    draw=ContextBorder,
    fill=ContextFill
},
function/.style={
    draw=FunctionBorder,
    fill=FunctionFill,
    rounded corners=1pt,
    minimum width=3.25cm,
    minimum height=0.36cm,
    inner xsep=2pt,
    inner ysep=1.2pt,
    align=center
}
]

\draw[panel]
    (-2.88,-0.72)
    rectangle
    (2.82,-3.06);

\draw[panel]
    (-2.88,-3.26)
    rectangle
    (2.82,-5.15);

\draw[panel]
    (-2.88,-5.35)
    rectangle
    (2.82,-6.80);

\draw[panel]
    (-2.88,-7.00)
    rectangle
    (2.82,-8.89);

\node[rootbox] (root) at (0,0)
    {Facade element};

\draw[spine]
    (root.south)
    -- (0,-0.48)
    -- (-3.28,-0.48)
    -- (-3.28,-7.28);

\node[
    structuregroup,
    anchor=west
] (structure) at (-2.65,-1.00)
    {Structure};

\node[structureclass] (wall)
    at (-1.42,-1.50)
    {Wall};

\node[structureclass] (roof)
    at (-1.42,-1.94)
    {Roof};

\node[structureclass] (column)
    at (-1.42,-2.38)
    {Column};

\node[structureclass] (arch)
    at (-1.42,-2.82)
    {Arch};

\node[function] (wallfun)
    at (1.08,-1.50)
    {LB, TE, VS};

\node[function] (rooffun)
    at (1.08,-1.94)
    {LB, TE, DR, FP};

\node[function] (columnfun)
    at (1.08,-2.38)
    {LB, AE};

\node[function] (archfun)
    at (1.08,-2.82)
    {LB, AE, IL, SU};

\draw[flow]
    (wall.east) -- (wallfun.west);

\draw[flow]
    (roof.east) -- (rooffun.west);

\draw[flow]
    (column.east) -- (columnfun.west);

\draw[flow]
    (arch.east) -- (archfun.west);

\node[
    openingsgroup,
    anchor=west
] (openings) at (-2.65,-3.54)
    {Openings};

\node[openingsclass] (window)
    at (-1.42,-4.04)
    {Window};

\node[openingsclass] (door)
    at (-1.42,-4.48)
    {Door};

\node[openingsclass] (blinds)
    at (-1.42,-4.92)
    {Blinds};

\node[function] (windowfun)
    at (1.08,-4.04)
    {TE, OP, SS, VE, NL};

\node[function] (doorfun)
    at (1.08,-4.48)
    {AC, OP, TE};

\node[function] (blindsfun)
    at (1.08,-4.92)
    {SS, OP, PS};

\draw[flow]
    (window.east) -- (windowfun.west);

\draw[flow]
    (door.east) -- (doorfun.west);

\draw[flow]
    (blinds.east) -- (blindsfun.west);

\node[
    appendagesgroup,
    anchor=west
] (appendages) at (-2.65,-5.63)
    {Appendages};

\node[appendagesclass] (balcony)
    at (-1.42,-6.13)
    {Balcony};

\node[appendagesclass] (stairs)
    at (-1.42,-6.57)
    {Stairs};

\node[function] (balconyfun)
    at (1.08,-6.13)
    {LB, AC, VS, FP};

\node[function] (stairsfun)
    at (1.08,-6.57)
    {CI, FP};

\draw[flow]
    (balcony.east) -- (balconyfun.west);

\draw[flow]
    (stairs.east) -- (stairsfun.west);

\node[
    contextgroup,
    anchor=west
] (context) at (-2.65,-7.28)
    {Context};

\node[contextclass] (terrain)
    at (-1.42,-7.78)
    {Terrain};

\node[contextclass] (vegetation)
    at (-1.42,-8.22)
    {Vegetation};

\node[contextclass] (other)
    at (-1.42,-8.66)
    {Other};

\node[function] (terrainfun)
    at (1.08,-7.78)
    {CI};

\node[function] (vegetationfun)
    at (1.08,-8.22)
    {AE};

\node[function] (otherfun)
    at (1.08,-8.66)
    {\textit{var.}};

\draw[flow]
    (terrain.east) -- (terrainfun.west);

\draw[flow]
    (vegetation.east) -- (vegetationfun.west);

\draw[flow]
    (other.east) -- (otherfun.west);

\draw[flow]
    (-3.28,-1.00)
    -- (structure.west);

\draw[flow]
    (-3.28,-3.54)
    -- (openings.west);

\draw[flow]
    (-3.28,-5.63)
    -- (appendages.west);

\draw[flow]
    (-3.28,-7.28)
    -- (context.west);

\end{tikzpicture}

\vspace{-1mm}

\caption{\textbf{Semantic-to-functional candidate mapping.} Each semantic class maps to a constrained candidate set of functional attributes under a multi-label formulation.
LB = load-bearing, TE = thermal envelope, DR = drainage,
FP = fall protection, AC = access, OP = operable,
CI = circulation, SS = solar shading, VE = ventilation,
NL = natural lighting, PS = privacy screening,
VS = vegetation support, AE = aesthetic,
IL = illumination, SU = surveillance, and
\textit{var.} denotes case-dependent attributes.}

\label{fig:sem2func_mapping}

\vspace{-2mm}
\end{figure}

\subsection{Functional Attribute Labeling}

\paragraph{Auto-suggestion.}
With semantic instances defined, we assigned candidate functional attributes using ontology- and rule-based heuristics established with reference to IFC~4.3 and the Getty Art \& Architecture Thesaurus \cite{ISO16739_2024,petersen1990art}.
Figure~\ref{fig:sem2func_mapping} maps each semantic class to a constrained multi-label candidate set, which provides the initial functional attribute suggestions. Each candidate carries an expert-set semantic prior $P(a\mid c)$, with per-class values given in the supplement.

\paragraph{Geometry- and Context-Guided Functional Refinement.}
Semantic labels provided functional priors but did not uniquely determine instance-level functionality.
Instances of the same semantic class could have different functions depending on their geometry and context, such as operable versus fixed windows or structural versus decorative columns.
Figure~\ref{fig:attr_examples} illustrates this within-class variation for windows, doors, and stairs.
Attributes such as \texttt{natural\_lighting} for windows were retained without refinement, while instance-dependent attributes were refined using instance geometry and graph-based contextual evidence, following Algorithm~\ref{alg:function_refinement}.

\begin{algorithm}[t]
\small
\DontPrintSemicolon
\caption{Geometry- and Context-Guided Functional Refinement}
\label{alg:function_refinement}

\KwIn{
instance $I$ with point set $\mathcal{P}_I$,\\
semantic class $c$, context graph $\mathcal{G}$
}
\KwOut{functional confidence vector $\mathbf{f}_I$}

Initialize semantic prior vector
$\mathbf{P}_I$ from $\mathbf{f}^{(0)}_c$\;

Initialize functional confidence vector
$\mathbf{f}_I \leftarrow \mathbf{P}_I$\;

Extract geometric descriptors from $\mathcal{P}_I$\;

Extract contextual relations from $\mathcal{G}$\;

\ForEach{candidate functional attribute $a$}{

\If{$a$ requires refinement}{
Compute geometric evidence $G_I(a)\in[0,1]$\;

Compute contextual evidence $C_I(a)\in[0,1]$\;
}
\Else{
$G_I(a)\leftarrow0.5$\;

$C_I(a)\leftarrow0.5$\;
}

$f_I(a)\leftarrow
\mathrm{Fuse}\!\left(P_I(a),G_I(a),C_I(a)\right)$\;
}

\Return $\mathbf{f}_I$\;
\end{algorithm}

For an instance $I$ and functional attribute $a$, we denote the semantic prior $P_I(a)$, geometric evidence $G_I(a)$, contextual evidence $C_I(a)$, and refined functional confidence $f_I(a)$.
Geometric evidence was derived from dimensions, depth, orientation, elevation, and ground relation, while contextual evidence captured proximity, overlap, attachment, and structural support relations.
The three sources were fused in the odds domain without class- or attribute-specific fusion weights.

\begin{equation}
\label{eq:functional_fusion}
\frac{f_I(a)}{1-f_I(a)}
=
\frac{P_I(a)}{1-P_I(a)}
\frac{G_I(a)}{1-G_I(a)}
\frac{C_I(a)}{1-C_I(a)}.
\end{equation}

Equation~\ref{eq:functional_fusion} assumes conditional independence between geometric and contextual evidence, which holds only approximately because contextual relations may partly depend on geometry and connectivity.
We therefore treat $f_I(a)$ as a functional confidence rather than a strictly calibrated posterior probability, and rely on the subsequent human validation to confirm or correct the suggested attributes rather than on these confidences alone.

For attributes not requiring refinement, we set $G_I(a)=C_I(a)=0.5$, corresponding to unit odds and leaving the semantic prior unchanged.
For attributes requiring refinement, evidence above $0.5$ strengthens the semantic prior, while evidence below $0.5$ weakens it.
$G$ and $C$ thus act as likelihood ratios relative to a neutral base rate of $0.5$, and an attribute is suggested as present when $f_I(a)\ge 0.5$.

\paragraph{Human Validation.}
The refined functional suggestions were subsequently reviewed by annotators using the associated ZEB scanner imagery and Google Street View~\cite{googlestreetview}.
During the review, each instance was highlighted in 3D together with its suggested functional attributes and corresponding confidence scores.
Annotators verified the suggested assignments, corrected ambiguous cases, and added attributes that were not reliably captured by the automatic refinement.
For example, annotators assessed window operability using visible geometric cues and the associated imagery, while they distinguished decorative columns from structural columns by revising their \texttt{load\_bearing} and \texttt{aesthetic} attributes.
The validated results were then used as the final instance-level functional annotations.

\paragraph{Label Verification and Audit.}
Each functional attribute follows IFC~4.3 and Getty AAT definitions and is read from the exterior LiDAR geometry and co-registered imagery, so a property such as \texttt{load\_bearing} is assessed visually, not verified against structural drawings. We audit the labels in two ways. First, across the two review rounds on ten buildings spanning the train, validation, and test splits, the rounds agree on 93.2\% of applicable attribute decisions. Second, in a test on two sampled buildings reviewed separately by two second-round annotators, each starting from the first-round labels, agreement reaches 99.0\% of decisions, with a Cohen's $\kappa$~\cite{cohen1960coefficient} of 0.93.

The final dataset associates each point with a semantic label and instance ID, while each instance is assigned a 15-dimensional functional attribute vector.
This combination of automatic refinement and human validation enables efficient instance-level functional annotation at dataset scale.
The dataset is partitioned at the building level into 18 training, 4 validation, and 4 test buildings, so no building contributes points to more than one split.
Point-wise semantic and instance annotations are provided together with separate instance-level functional attribute files. The supplementary material provides additional acquisition and annotation details, per-building review and instance-clustering statistics, implementation settings, extended qualitative comparisons, and the complete cross-dataset evaluation matrix.

\section{Benchmark Evaluation}
\label{sec:benchmark}

We establish a comprehensive benchmark for urban facade parsing across three tasks: (i) \textit{semantic segmentation} of facade points, (ii) \textit{instance segmentation} of individual facade elements, and (iii) \textit{functional attribute labeling} at the instance level. Urban facades combine strong geometric regularities, such as coplanar windows and repeated balconies, with substantial functional diversity, including access, ventilation, and shading. While most existing 3D benchmarks focus on semantic labeling, instance-level and functional reasoning remain comparatively underexplored for large-scale outdoor point clouds. Sections~\ref{subsec:sem_models} and~\ref{subsec:insseg} benchmark seven semantic architectures and three learned instance methods. Because the learned instance methods must infer object boundaries without the true semantic class of each point, we additionally report a geometry-based clustering baseline given ground-truth semantics; the two are reported in separate tables rather than merged into one leaderboard. Section~\ref{subsec:crossdomain} measures cross-dataset transfer, and Section~\ref{subsec:func_eval} examines functional attributes.
\begin{table}[h]
\centering
\caption{\textbf{Semantic segmentation overall metrics.} Best per column in \textbf{bold}.}
\label{tab:semantic}
\resizebox{\columnwidth}{!}{%
\begin{tabular}{lccccc}
\toprule
\textbf{Method} & \textbf{mIoU} & \textbf{OA} & \textbf{mP} & \textbf{mR} & \textbf{mF1} \\
\midrule
PointNet~\cite{qi2017pointnet}         & 26.30 & 75.63 & 39.78 & 32.94 & 33.32 \\
PointNet++~\cite{qi2017pointnetpp}     & 29.19 & 78.54 & 39.14 & 34.76 & 35.24 \\
DGCNN~\cite{wang2019dynamic}           & 15.37 & 71.94 & 25.70 & 18.93 & 18.93 \\
KPConv~\cite{thomas2019kpconv}         & \textbf{47.26} & \textbf{86.64} & \textbf{67.06} & \textbf{53.54} & \textbf{57.15} \\
RandLA-Net~\cite{hu2020randla}         & 16.94 & 72.64 & 24.92 & 21.43 & 21.39 \\
Point Transformer v2~\cite{wu2022ptv2} & 20.65 & 77.94 & 33.00 & 27.53 & 24.95 \\
Point Transformer v3~\cite{wu2024ptv3} & 28.18 & 74.76 & 45.11 & 33.66 & 35.66 \\
\bottomrule
\end{tabular}
}
\vspace{-10pt}
\end{table}

\begin{table*}[t]
\centering
\caption{\textbf{Per-class IoU (\%).} Classes ordered from most to least frequent. Best per class in \textbf{bold}.}
\label{tab:semantic_perclass}
\resizebox{\textwidth}{!}{%
\begin{tabular}{lcccccccccccc}
\toprule
\textbf{Method} & \textbf{Wall} & \textbf{Terrain} & \textbf{Vegetation} & \textbf{Window} &
\textbf{Blinds} & \textbf{Other} & \textbf{Column} & \textbf{Balcony} &
\textbf{Roof} & \textbf{Stairs} & \textbf{Door} & \textbf{Arch} \\
\midrule
PointNet~\cite{qi2017pointnet}         & 69.31 & 73.46 & 81.27 & 20.45 & 4.27 & 22.19 & 2.20 & 0.00 & 0.13 & 41.14 & 1.16 & 0.00 \\
PointNet++~\cite{qi2017pointnetpp}     & 72.47 & 83.57 & 84.88 & 21.99 & 6.39 & 16.25 & 1.02 & 0.00 & 0.02 & 63.73 & 0.00 & 0.00 \\
DGCNN~\cite{wang2019dynamic}           & 63.53 & 83.10 & 33.25 & 4.21 & 0.00 & 0.34 & 0.00 & 0.00 & 0.00 & 0.00 & 0.00 & 0.00 \\
KPConv~\cite{thomas2019kpconv}         & \textbf{80.37} & \textbf{91.59} & \textbf{93.94} & \textbf{42.05} & \textbf{42.05} & \textbf{53.73} & \textbf{35.80} & 24.71 & 0.00 & \textbf{83.43} & \textbf{19.47} & \textbf{0.01} \\
RandLA-Net~\cite{hu2020randla}         & 67.98 & 82.04 & 30.05 & 3.01 & 0.00 & 0.00 & 0.00 & 0.00 & 20.14 & 0.00 & 0.00 & 0.00 \\
Point Transformer v2~\cite{wu2022ptv2} & 77.74 & 80.16 & 58.81 & 29.14 & 0.00 & 1.88 & 0.00 & 0.05 & 0.00 & 0.00 & 0.00 & 0.00 \\
Point Transformer v3~\cite{wu2024ptv3} & 65.98 & 83.91 & 83.41 & 6.62 & 0.00 & 28.20 & 10.44 & \textbf{36.65} & \textbf{22.89} & 0.00 & 0.00 & 0.00 \\
\bottomrule
\end{tabular}%
}
\end{table*}
\begin{figure*}[t]
\centering
\includegraphics[width=\linewidth]{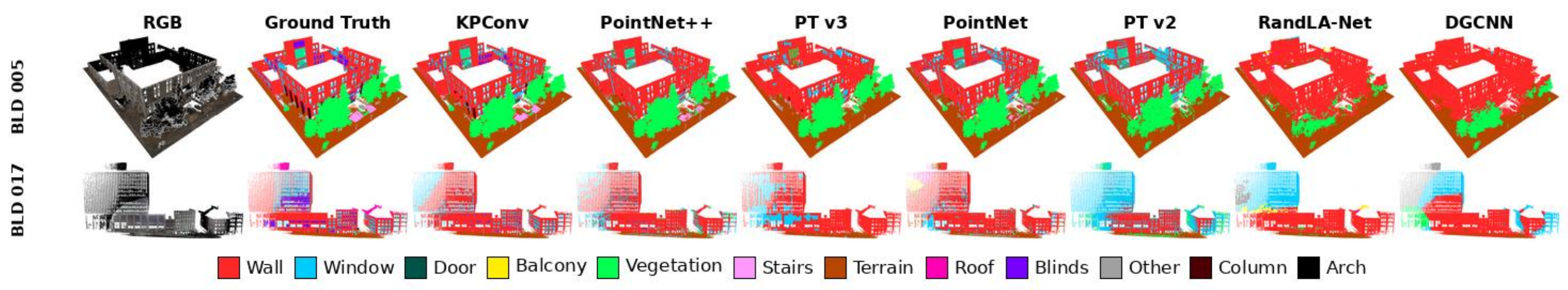}
\caption{\textbf{Qualitative semantic segmentation on two test buildings.} Columns show the
RGB scan, ground truth, and predictions from all seven backbones ordered left-to-right by
decreasing mIoU; rows show a low-rise and a high-rise building. KPConv and PointNet++ recover
windows, vegetation, and stairs close to ground truth, while RandLA-Net and DGCNN wash most of
the facade into wall and terrain.}
\label{fig:qual_semantic}
\end{figure*}

All models are trained on the training split and evaluated on the held-out test split defined in Section~\ref{sec:dataset}; the training, validation, and test splits contain 343.7M, 68.4M, and 79.7M points, respectively. To ensure fair cross-architecture comparison, the evaluation constraint is 100\% point coverage before any metric is computed rather than one prescribed sampling mechanism. Six of the seven models meet it through a shared possibility-guided KNN-crop procedure: crops are sampled iteratively around the least-covered point via a cKD-tree over the full building, and per-point softmax probabilities from overlapping crops are averaged and argmax-decoded. Crop sizes follow each model's native input scale, ranging from 4{,}096 to 120k points. KPConv instead uses the potential-guided spherical sampling of its reference implementation, which provides the same coverage guarantee; its radius-neighbourhood input pipeline cannot consume fixed-size crops without reimplementation. No model uses test-time augmentation. All experiments share the same 7-dimensional input, $[x,y,z,r,g,b,\text{intensity}]$: XYZ mean-centred per building (metric scale preserved), RGB normalised to $[0,1]$, and intensity normalised via the training-split $p_1$/$p_{99}$ percentiles.

\subsection{Benchmarking Task: Semantic Segmentation}
\label{subsec:sem_models}
We benchmark seven architectures spanning point-based, graph-based, convolutional, and transformer-based paradigms: PointNet~\cite{qi2017pointnet}, PointNet++~\cite{qi2017pointnetpp}, DGCNN~\cite{wang2019dynamic}, KPConv~\cite{thomas2019kpconv}, RandLA-Net~\cite{hu2020randla}, Point Transformer v2~\cite{wu2022ptv2}, and Point Transformer v3~\cite{wu2024ptv3}. As shown in Table~\ref{tab:semantic}, KPConv achieves the highest performance with 47.26\% mIoU. Notably, PointNet and PointNet++ outperform Point Transformer v2, while DGCNN ranks last, differing from the ordering commonly observed on indoor benchmarks such as S3DIS~\cite{armeni2016cvpr}. The per-class results in Table~\ref{tab:semantic_perclass} reveal a strong dependence on class frequency: performance is highest for the dominant wall and terrain classes and degrades substantially for rare categories. In particular, arch, which accounts for only 0.3\% of the points, is recovered with negligible IoU across all models. Figure~\ref{fig:qual_semantic} further compares predictions from all seven backbones on representative low- and high-rise test buildings.

This consistent dependence on class frequency suggests that class imbalance is a major bottleneck in the benchmark and may partly explain why increasing architectural sophistication does not translate directly into higher overall performance. Similar challenges have been reported for facade point-cloud benchmarks such as TUM-FACADE~\cite{wysockiTUMFACADE} and ZAHA~\cite{wysocki2024zahaintroducinglevelfacade}. Improving recognition of rare facade elements may therefore require imbalance-aware strategies, such as class-balanced re-weighting~\cite{cui2019classbalanced} or focal loss~\cite{lin2017focal}, in addition to advances in model architecture.

\begin{table}[h]
\centering
\caption{\textbf{Learned instance segmentation} on the held-out test set, with one training run per method and the checkpoint selected on validation. All three methods are trained from raw point features; no semantic labels are supplied as input.}
\label{tab:instance}
\begin{tabular}{lccc}
\toprule
\textbf{Method} & \textbf{mAP} & \textbf{AP50} & \textbf{AP25} \\
\midrule
PointGroup~\cite{jiang2020pointgroup} & 0.0145 & 0.0427 & 0.0955 \\
ISBNet~\cite{ngo2023isbnet}           & 0.0422 & 0.0883 & 0.2276 \\
SoftGroup~\cite{vu2022softgroup}      & \textbf{0.2032} & \textbf{0.2923} & \textbf{0.4044} \\
\bottomrule
\end{tabular}
\vspace{-10pt}
\end{table}

\subsection{Benchmarking Task: Instance Segmentation}
\label{subsec:insseg}
Three learned instance segmentation methods are trained from scratch on raw point features, without ground-truth semantic labels: PointGroup~\cite{jiang2020pointgroup}, which groups points using original and offset-shifted coordinates; ISBNet~\cite{ngo2023isbnet}, which combines instance-aware sampling with box-aware dynamic convolution; and SoftGroup~\cite{vu2022softgroup}, which performs soft semantic grouping followed by top-down refinement. SoftGroup leads with 0.29 AP50, ahead of ISBNet (0.09) and PointGroup (0.04) (Table~\ref{tab:instance}). Instance AP here depends heavily on the pipeline, not only the data. Scoring subsampled predictions against full-resolution ground truth caps mask IoU and gives near-zero AP: ground-truth masks reach only 0.31 AP50 on one validation building, against 0.91 with masks propagated to every point. A 0.9\,m grouping radius also chains neighbouring elements. We therefore propagate masks to every point before scoring and group PointGroup and SoftGroup at 7.5\,cm.

The remaining difficulty is class-specific. PointGroup over-fragments: the 52 trees of one test building receive 358 confident vegetation predictions, which outrank correct predictions elsewhere in pooled AP. ISBNet scores well only on balconies (0.54 AP50). Column and stairs, with only 14 and 5 test instances, supply about 70\% of SoftGroup's mAP. Windows, the most numerous class, are easier to locate than to delineate: SoftGroup's window AP falls from 0.56 at IoU 0.25 to 0.22 at IoU 0.50.

Table~\ref{tab:instance_oracle} provides a complementary diagnostic by evaluating geometry-based clustering given ground-truth semantic labels, thereby isolating instance separation from semantic recognition. HDBSCAN~\cite{campello2013density} and DBSCAN~\cite{ester1996density} lead with 0.34 and 0.27 AP50, ahead of MeanShift~\cite{comaniciu2002mean} and KMeans~\cite{lloyd1982least}. Well-separated elements such as balconies and columns are recovered well under HDBSCAN (0.62 and 0.66 AP50), but windows are not (0.05), because density clustering merges neighbouring panes.
\begin{table}[h]
\centering
\caption{\textbf{Geometry-based clustering given ground-truth semantics}, test set. An oracle-semantic clustering baseline: each point's class is given, so it is not directly comparable to Table~\ref{tab:instance}, which receives no semantic information.}
\label{tab:instance_oracle}
\begin{tabular}{lccc}
\toprule
\textbf{Method} & \textbf{mAP} & \textbf{AP50} & \textbf{AP25} \\
\midrule
DBSCAN~\cite{ester1996density}    & 0.2063 & 0.2725 & 0.3574 \\
HDBSCAN~\cite{campello2013density} & \textbf{0.2533} & \textbf{0.3430} & \textbf{0.4521} \\
MeanShift~\cite{comaniciu2002mean} & 0.0310 & 0.0871 & 0.1059 \\
KMeans~\cite{lloyd1982least}       & 0.0201 & 0.0615 & 0.1793 \\
\bottomrule
\end{tabular}
\vspace{-10pt}
\end{table}

These results locate the difficulty in dense, repeated elements such as windows and blinds rather than in every class, while rare classes make macro-averaged AP volatile on this test set. They motivate methods that explicitly account for repetition, facade-scale geometry, and spatial regularity.

\subsection{Cross-Dataset Generalization}
\label{subsec:crossdomain}

To test how well the proposed benchmark generalises beyond its own facades, we compare it against ZAHA~\cite{wysocki2024zahaintroducinglevelfacade}, the closest published facade benchmark, remapping both datasets to a shared 11-class taxonomy (vegetation, absent from ZAHA, is dropped) and training with xyz-only input for a feature-neutral comparison. All seven architectures follow the same protocol as Section~\ref{subsec:sem_models}.

\begin{figure}[t]
\centering
\includegraphics[width=\linewidth]{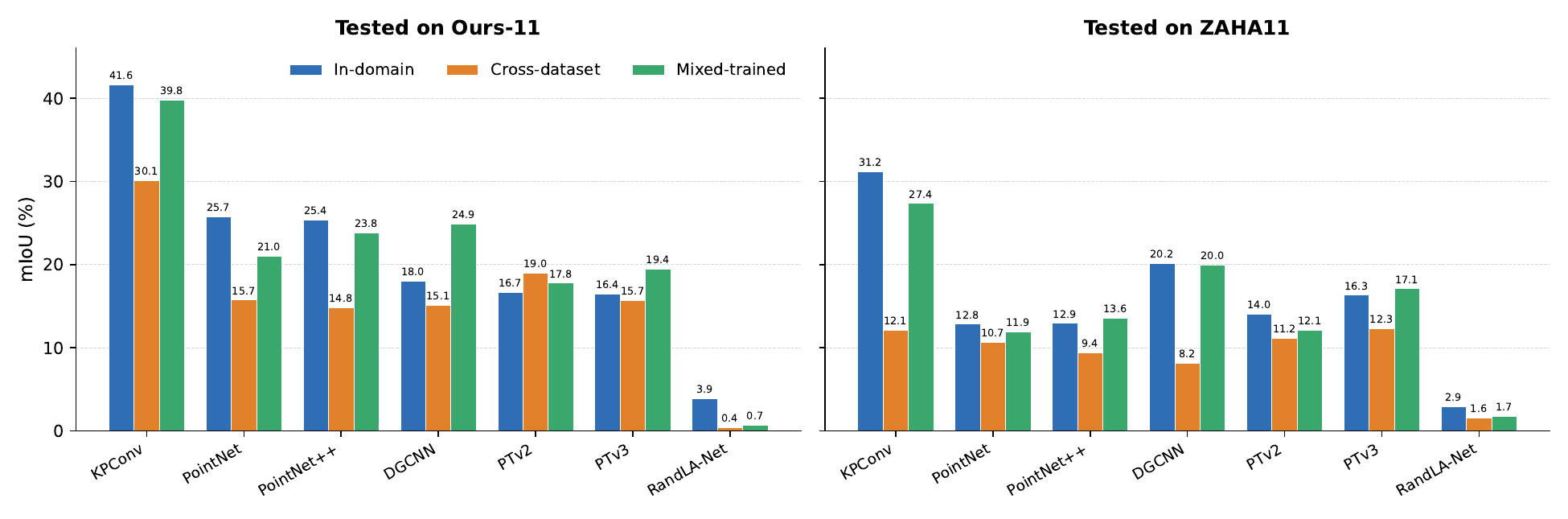}
\caption{\textbf{Cross-dataset mIoU by test domain.} For each architecture: in-domain, cross-dataset, and mixed-trained (HFX11 + ZAHA11) mIoU, evaluated on HFX11 (left) and ZAHA11 (right).}
\label{fig:cross_domain}
\vspace{-10pt}
\end{figure}

KPConv leads every setting in Figure~\ref{fig:cross_domain}, but its cross-dataset transfer is markedly asymmetric: training on HFX3D and testing on ZAHA loses 29.5 percentage points, while the reverse loses only one. This asymmetry holds across architectures: every model tested on ZAHA does worse when trained on HFX3D instead, a uniform domain penalty absent in the other direction. Tested on HFX3D, cross-dataset training is architecture-dependent rather than uniformly harmful: Point Transformer v2 actually improves when trained on ZAHA instead of HFX3D itself, and Point Transformer v3 is essentially unaffected, while the remaining architectures lose substantially.

This asymmetry likely reflects several factors rather than a single cause: HFX3D's broader environmental and architectural variability, together with the label-space harmonization (shared 11-class taxonomy, dropped vegetation and colour) and annotation-convention differences between the datasets. We therefore report it as an observation rather than evidence of superior robustness.

Mixed training tells a similar story: combining both datasets helps DGCNN and Point Transformer v3 on HFX3D, but hurts KPConv, PointNet, and RandLA-Net, with a smaller effect either way on ZAHA.

\begin{table}[h]
\caption{\textbf{Evidence ablation for functional attribute labeling.}
Semantic prior ($S$), geometric evidence ($G$), and graph context ($C$) combine in log-odds space.
An ablated source is fixed to $0.5$.
Macro-F1 (mF1) and exact set match (ESM) are reported on all evaluated test instances and on the class-ambiguous subset.
The learned MLP row is the mean over 10 seeds.
}
\centering
\small
\setlength{\tabcolsep}{4pt}
\begin{tabular}{lcccc}
\toprule
& \multicolumn{2}{c}{All} & \multicolumn{2}{c}{Ambiguous} \\
\cmidrule(lr){2-3}\cmidrule(lr){4-5}
Evidence & mF1 & ESM & mF1 & ESM \\
\midrule
$S$ only          & 0.6367          & \textbf{0.6905} & 0.4808          & 0.5378 \\
$G$ only          & \textbf{0.8071} & 0.5880          & \textbf{0.7039} & 0.5280 \\
$C$ only          & 0.7449          & 0.5420          & 0.6746          & 0.4958 \\
$S{+}G$           & 0.7458          & 0.6888          & 0.5928          & \textbf{0.5434} \\
$S{+}C$           & 0.7145          & 0.6676          & 0.6072          & 0.5224 \\
$S{+}G{+}C$       & 0.7691          & 0.6799          & 0.6606          & \textbf{0.5434} \\
Learned MLP       & 0.7304          & 0.5590          & 0.6354          & 0.4141 \\
\bottomrule
\end{tabular}
\vspace{-10pt}
\label{tab:attr_ablation}
\end{table}

\subsection{Functional Attribute Labeling}
\label{subsec:func_eval}

Functional attributes are assigned by the pipeline of Section~\ref{sec:dataset}, which fuses a semantic prior $S$, geometric evidence $G$, and graph context $C$.
We ablate these three sources against the human-validated labels to quantify how much of the within-class functional variation each recovers.
As a learned baseline, a small MLP is trained on 21 per-instance metric descriptors that summarise each element's size, shape, elevation, orientation, and point density, with the checkpoint selected on validation.
We report results on two subsets of the test split. 
\emph{All} comprises every annotated instance. 
\emph{Ambiguous} restricts to instances whose semantic class
does not determine the attribute set. 
For a class $c$ and functional attribute
$a$, let $\mathcal{I}_c$ denote the instances of class $c$ in the training
buildings and $A_i$ the validated attribute set of instance $i$. 
We define the ground-truth attribute rate as in Equation \ref{eq:attr_rate}.
\begin{equation}
r(c,a) \;=\; \frac{\bigl|\{\, i \in \mathcal{I}_c \;:\; a \in A_i \,\}\bigr|}
                  {|\mathcal{I}_c|} ,
\label{eq:attr_rate}
\end{equation}
and the pair $(c,a)$ is ambiguous if $\tau \le r(c,a) \le 1-\tau$, with
$\tau{=}0.05$. 
Rates near $0$ or $1$ indicate that class membership already
fixes the attribute; intermediate rates indicate functional variation within the
class. 
An instance enters \emph{Ambiguous} if its class admits at least one
ambiguous pair, since attributes are predicted jointly and exact set match is only defined over the full set.
The training split contains 16 ambiguous pairs across seven semantic classes.
The resulting \emph{All} and \emph{Ambiguous} test subsets contain 1,131 and 714 evaluated instances, respectively.\\

Functional attribute labeling remains challenging for several reasons.
Because semantic class does not fix instance-level function, $S$ only reaches an mF1 of 0.4808 on \emph{Ambiguous} (Table~\ref{tab:attr_ablation}).
Geometry and context provide instance-level evidence, but the partial surface coverage and uneven point density of facade scans restrict the reliability of both sources.
Limited point coverage may omit handles, opening seams, or articulated components that distinguish an operable window from a fixed window.
Occlusion from neighbouring components and vegetation may obscure structural junctions and prevent the context graph from recovering attachment and support relations.
The additional evidence therefore improves the \textit{Ambiguous }mF1 to 0.7039 with $G$ only and 0.6746 with $C$ only, but neither source resolves the functional ambiguity completely.
Geometry and context may also support different assignments for the same attribute.
For example, the vertical shape and ground connection of a wall may support \texttt{load\_bearing}, while an incomplete connection to the upper structure may provide weak contextual support for the same attribute.
Such incomplete or opposing evidence limits $S{+}G{+}C$ to an Ambiguous mF1 of 0.6606.
The highest Ambiguous ESM remains 0.5434 because one incorrect applicable attribute invalidates the complete instance-level attribute set.
Rare attributes such as \texttt{drainage} and \texttt{vegetation\_support} create an additional difficulty because their limited positive samples restrict reliable functional assignment.

\section{Conclusion}
\label{sec:conclusion}

We introduced HFX3D, a large-scale LiDAR facade benchmark with per-instance functional annotations, spanning approximately 492 million points across 26 diverse urban buildings in Halifax, Canada. Beyond standard semantic labels, the benchmark provides 15 functional attributes per instance, derived through architectural ontologies, geometry-guided heuristics, and human validation. We define three tasks of increasing complexity: point-level semantic segmentation, object-level instance grouping, and instance-level functional attribute labeling. We benchmark these tasks across seven semantic architectures, three learned instance methods plus four geometry-clustering baselines, and an evidence ablation for functional attributes. Semantic segmentation remains difficult under class imbalance and glass-surface sparsity. Instance AP stays low, concentrated in a few rare, well-separated classes, while dense repeated windows resist delineation. Functional labeling is hardest, especially for ambiguous attributes such as \texttt{operable} and \texttt{aesthetic} and rare ones such as \texttt{drainage} and \texttt{vegetation\_support}.

The benchmark establishes a new level of annotation richness for 3D urban understanding and opens directions at the intersection of point cloud learning, architectural knowledge representation, and BIM. Future work includes extending functional annotations to interior scans, incorporating complementary sensing modalities such as thermal infrared, and developing architectures that jointly optimise semantic, instance, and functional reasoning through shared functional-geometric representations.

\section*{Acknowledgments}
This work was supported by NSERC ( RGPIN: \#05092-2019) and CFI (JELF \#42358). We thank Prachi Kudeshia, Jack Greenlaw, Avery Cao, Khushal Das, and Mirza Baig for data collection and annotation.

   {\small
   \bibliographystyle{ieeenat_fullname}
   \bibliography{main,graz50}

\begin{thebibliography}{49}
\providecommand{\natexlab}[1]{#1}
\providecommand{\url}[1]{\texttt{#1}}
\expandafter\ifx\csname urlstyle\endcsname\relax
  \providecommand{\doi}[1]{doi: #1}\else
  \providecommand{\doi}{doi: \begingroup \urlstyle{rm}\Url}\fi

\bibitem[Armeni et~al.(2016)Armeni, Sener, Zamir, Jiang, Brilakis, Fischer, and Savarese]{armeni2016cvpr}
Iro Armeni, Ozan Sener, Amir~R. Zamir, Helen Jiang, Ioannis Brilakis, Martin Fischer, and Silvio Savarese.
\newblock 3d semantic parsing of large-scale indoor spaces.
\newblock In \emph{Proceedings of the IEEE Conference on Computer Vision and Pattern Recognition (CVPR)}, 2016.

\bibitem[Behley et~al.(2019)Behley, Garbade, Milioto, Quenzel, Behnke, Stachniss, and Gall]{behley2019semantickitti}
Jens Behley, Martin Garbade, Andres Milioto, Jan Quenzel, Sven Behnke, Cyrill Stachniss, and Juergen Gall.
\newblock {SemanticKITTI}: A dataset for semantic scene understanding of lidar sequences.
\newblock In \emph{Proceedings of the IEEE/CVF International Conference on Computer Vision (ICCV)}, pages 9297--9307, 2019.

\bibitem[Biljecki et~al.(2015)Biljecki, Stoter, Ledoux, Zlatanova, and {\c{C}}{\"o}ltekin]{biljecki2015applications}
Filip Biljecki, Jantien Stoter, Hugo Ledoux, Sisi Zlatanova, and Arzu {\c{C}}{\"o}ltekin.
\newblock Applications of 3d city models: State of the art review.
\newblock \emph{ISPRS International Journal of Geo-Information}, 4\penalty0 (4):\penalty0 2842--2889, 2015.

\bibitem[Campello et~al.(2013)Campello, Moulavi, and Sander]{campello2013density}
Ricardo J. G.~B. Campello, Davoud Moulavi, and J{\"o}rg Sander.
\newblock Density-based clustering based on hierarchical density estimates.
\newblock In \emph{Advances in Knowledge Discovery and Data Mining (PAKDD)}, pages 160--172. Springer, 2013.

\bibitem[Chen et~al.(2026)Chen, Li, Han, Feng, Chen, and Wang]{chen2026cityfacade}
Yiping Chen, Jonathan Li, Ting Han, Huifang Feng, Jun Chen, and Cheng Wang.
\newblock {City-Facade}: A city-level large-scale point cloud building facade dataset for semantic \& instance segmentation.
\newblock \emph{ISPRS Journal of Photogrammetry and Remote Sensing}, 2026.

\bibitem[Cheung and Lee(2024)]{cheung2024arch2sdatasetbenchmarkchallenges}
Ka~Lung Cheung and Chi~Chung Lee.
\newblock Arch2s: Dataset, benchmark and challenges for learning exterior architectural structures from point clouds, 2024.

\bibitem[Cohen(1960)]{cohen1960coefficient}
Jacob Cohen.
\newblock A coefficient of agreement for nominal scales.
\newblock \emph{Educational and Psychological Measurement}, 20\penalty0 (1):\penalty0 37--46, 1960.

\bibitem[Comaniciu and Meer(2002)]{comaniciu2002mean}
Dorin Comaniciu and Peter Meer.
\newblock Mean shift: A robust approach toward feature space analysis.
\newblock \emph{IEEE Transactions on Pattern Analysis and Machine Intelligence}, 24\penalty0 (5):\penalty0 603--619, 2002.

\bibitem[Cui et~al.(2019)Cui, Jia, Lin, Song, and Belongie]{cui2019classbalanced}
Yin Cui, Menglin Jia, Tsung-Yi Lin, Yang Song, and Serge Belongie.
\newblock Class-balanced loss based on effective number of samples.
\newblock In \emph{Proceedings of the IEEE/CVF Conference on Computer Vision and Pattern Recognition (CVPR)}, pages 9268--9277, 2019.

\bibitem[Dai et~al.(2017)Dai, Chang, Savva, Halber, Funkhouser, and Nie{\ss}ner]{dai2017scannet}
Angela Dai, Angel~X. Chang, Manolis Savva, Maciej Halber, Thomas Funkhouser, and Matthias Nie{\ss}ner.
\newblock Scannet: Richly-annotated 3d reconstructions of indoor scenes.
\newblock In \emph{Proc. Computer Vision and Pattern Recognition (CVPR), IEEE}, 2017.

\bibitem[{Datacluster Labs}(2021)]{datacluster_stairs}
{Datacluster Labs}.
\newblock Stairs image dataset.
\newblock \url{https://github.com/datacluster-labs/Stairs-Image-Dataset}, 2021.
\newblock Accessed: 2026-08-27.

\bibitem[Ester et~al.(1996)Ester, Kriegel, Sander, and Xu]{ester1996density}
Martin Ester, Hans-Peter Kriegel, J{\"o}rg Sander, and Xiaowei Xu.
\newblock A density-based algorithm for discovering clusters in large spatial databases with noise.
\newblock In \emph{Proceedings of the Second International Conference on Knowledge Discovery and Data Mining (KDD)}, pages 226--231, 1996.

\bibitem[{FARO Technologies, Inc.}(2024)]{faroconnect}
{FARO Technologies, Inc.}
\newblock {FARO Connect} (formerly {GeoSLAM Connect}) [software], 2024.

\bibitem[{GeoSLAM (FARO Technologies, Inc.)}(2024)]{zebhorizon}
{GeoSLAM (FARO Technologies, Inc.)}.
\newblock {ZEB Horizon} mobile mapping system [hardware], 2024.

\bibitem[Girardeau-Montaut(2011)]{girardeau2011cloudcompare}
Daniel Girardeau-Montaut.
\newblock {CloudCompare}, a 3{D} point cloud and mesh processing free software.
\newblock Technical report, EDF Research and Development, T{\'e}l{\'e}com ParisTech, 2011.

\bibitem[{Google}()]{googlestreetview}
{Google}.
\newblock {Google Street View}.

\bibitem[Gr{\"o}ger et~al.(2012)Gr{\"o}ger, Kolbe, Nagel, and H{\"a}fele]{groger2012citygml}
Gerhard Gr{\"o}ger, Thomas~H Kolbe, Claus Nagel, and Karl-Heinz H{\"a}fele.
\newblock Citygml--interoperable semantic 3d city models.
\newblock \emph{ISPRS Journal of Photogrammetry and Remote Sensing}, 71:\penalty0 12--33, 2012.

\bibitem[Hackel et~al.(2017)Hackel, Savinov, Ladicky, Wegner, Schindler, and Pollefeys]{hackel2017isprs}
Timo Hackel, N. Savinov, L. Ladicky, Jan~D. Wegner, K. Schindler, and M. Pollefeys.
\newblock {SEMANTIC3D.NET: A new large-scale point cloud classification benchmark}.
\newblock In \emph{ISPRS Annals of the Photogrammetry, Remote Sensing and Spatial Information Sciences}, pages 91--98, 2017.

\bibitem[Hu et~al.(2020)Hu, Yang, Xie, Rosa, Guo, Wang, Trigoni, and Markham]{hu2020randla}
Qingyong Hu, Bo Yang, Linhai Xie, Stefano Rosa, Yulan Guo, Zhiqiang Wang, Niki Trigoni, and Andrew Markham.
\newblock Randla-net: Efficient semantic segmentation of large-scale point clouds.
\newblock In \emph{Proceedings of the IEEE/CVF Conference on Computer Vision and Pattern Recognition (CVPR)}, pages 11108--11117, 2020.

\bibitem[Hu et~al.(2022)Hu, Yang, Khalid, Xiao, Trigoni, and Markham]{hu2021sensaturban}
Qingyong Hu, Bo Yang, Sheikh Khalid, Wen Xiao, Niki Trigoni, and Andrew Markham.
\newblock Sensaturban: Learning semantics from urban-scale photogrammetric point clouds.
\newblock \emph{International Journal of Computer Vision}, 130:\penalty0 316--343, 2022.

\bibitem[{International Organization for Standardization}(2024)]{ISO16739_2024}
{International Organization for Standardization}.
\newblock Industry foundation classes ({IFC}) for data sharing in the construction and facility management industries---part 1: Data schema, 2024.
\newblock {ISO} 16739-1:2024.

\bibitem[Jiang et~al.(2020)Jiang, Zhao, Shi, Liu, Fu, and Jia]{jiang2020pointgroup}
Li Jiang, Hengshuang Zhao, Shaoshuai Shi, Shu Liu, Chi-Wing Fu, and Jiaya Jia.
\newblock Pointgroup: Dual-set point grouping for 3d instance segmentation.
\newblock In \emph{CVPR}, pages 4867--4876, 2020.

\bibitem[Kolbe et~al.(2005)Kolbe, Gr{\"o}ger, and Pl{\"u}mer]{kolbe2005citygml}
Thomas~H Kolbe, Gerhard Gr{\"o}ger, and Lutz Pl{\"u}mer.
\newblock Citygml: Interoperable access to 3d city models.
\newblock \emph{Geo-information for disaster management}, pages 883--899, 2005.

\bibitem[Kor{\v c} and F{\"o}rstner(2009)]{korc2009etrims}
Filip Kor{\v c} and Wolfgang F{\"o}rstner.
\newblock etrims image database for interpreting images of man-made scenes.
\newblock Technical Report TR-IGG-P-2009-01, Dept. of Photogrammetry, University of Bonn, 2009.

\bibitem[Liao et~al.(2022)Liao, Xie, and Geiger]{Liao2022PAMI}
Yiyi Liao, Jun Xie, and Andreas Geiger.
\newblock {KITTI}-360: A novel dataset and benchmarks for urban scene understanding in 2d and 3d.
\newblock \emph{Pattern Analysis and Machine Intelligence (PAMI)}, 2022.

\bibitem[Lin et~al.(2017)Lin, Goyal, Girshick, He, and Doll{\'a}r]{lin2017focal}
Tsung-Yi Lin, Priya Goyal, Ross Girshick, Kaiming He, and Piotr Doll{\'a}r.
\newblock Focal loss for dense object detection.
\newblock In \emph{Proceedings of the IEEE International Conference on Computer Vision (ICCV)}, pages 2980--2988, 2017.

\bibitem[Lloyd(1982)]{lloyd1982least}
Stuart~P. Lloyd.
\newblock Least squares quantization in {PCM}.
\newblock \emph{IEEE Transactions on Information Theory}, 28\penalty0 (2):\penalty0 129--137, 1982.

\bibitem[Martinovic and Van~Gool(2013)]{martinovic2015bayesian}
Andelo Martinovic and Luc Van~Gool.
\newblock Bayesian grammar learning for inverse procedural modeling.
\newblock In \emph{Proceedings of the IEEE Conference on Computer Vision and Pattern Recognition (CVPR)}, pages 201--208, 2013.

\bibitem[Matrone et~al.(2020)Matrone, Lingua, Pierdicca, Malinverni, Paolanti, Grilli, Remondino, Murtiyoso, and Landes]{matrone2020arch}
F. Matrone, A. Lingua, R. Pierdicca, E.~S. Malinverni, M. Paolanti, E. Grilli, F. Remondino, A. Murtiyoso, and T. Landes.
\newblock A benchmark for large-scale heritage point cloud semantic segmentation.
\newblock In \emph{Proceedings of the International Archives of the Photogrammetry, Remote Sensing and Spatial Information Sciences}, pages 1419--1426, 2020.

\bibitem[Musialski et~al.(2013)Musialski, Wonka, Aliaga, Wimmer, Van~Gool, and Purgathofer]{musialski2013survey}
Przemyslaw Musialski, Peter Wonka, Daniel~G Aliaga, Michael Wimmer, Luc Van~Gool, and Werner Purgathofer.
\newblock A survey of urban reconstruction.
\newblock \emph{Computer graphics forum}, 32\penalty0 (6):\penalty0 146--177, 2013.

\bibitem[Ngo et~al.(2023)Ngo, Hua, and Nguyen]{ngo2023isbnet}
Tuan~Duc Ngo, Binh-Son Hua, and Khoi Nguyen.
\newblock {ISBNet}: a 3{D} point cloud instance segmentation network with instance-aware sampling and box-aware dynamic convolution.
\newblock In \emph{Proceedings of the IEEE/CVF Conference on Computer Vision and Pattern Recognition (CVPR)}, 2023.

\bibitem[Petersen(1990)]{petersen1990art}
Toni Petersen, editor.
\newblock \emph{Art \& Architecture Thesaurus}.
\newblock Oxford University Press, New York, 1990.

\bibitem[Qi et~al.(2017{\natexlab{a}})Qi, Su, Mo, and Guibas]{qi2017pointnet}
Charles~R Qi, Hao Su, Kaichun Mo, and Leonidas~J Guibas.
\newblock Pointnet: Deep learning on point sets for 3d classification and segmentation.
\newblock In \emph{CVPR}, pages 652--660, 2017{\natexlab{a}}.

\bibitem[Qi et~al.(2017{\natexlab{b}})Qi, Yi, Su, and Guibas]{qi2017pointnetpp}
Charles~Ruizhongtai Qi, Li Yi, Hao Su, and Leonidas~J Guibas.
\newblock Pointnet++: Deep hierarchical feature learning on point sets in a metric space.
\newblock In \emph{NeurIPS}, pages 5099--5108, 2017{\natexlab{b}}.

\bibitem[Riemenschneider et~al.(2012)Riemenschneider, Krispel, Thaller, Donoser, Havemann, Fellner, and Bischof]{riemenschneider2012irregular}
Hayko Riemenschneider, Ulrich Krispel, Wolfgang Thaller, Michael Donoser, Sven Havemann, Dieter Fellner, and Horst Bischof.
\newblock Irregular lattices for complex shape grammar facade parsing.
\newblock In \emph{Proceedings of the IEEE Conference on Computer Vision and Pattern Recognition (CVPR)}, pages 1640--1647, 2012.

\bibitem[Roynard et~al.(2017)Roynard, Deschaud, and Goulette]{DBLP:journals/corr/abs-1712-00032}
Xavier Roynard, Jean{-}Emmanuel Deschaud, and Fran{\c{c}}ois Goulette.
\newblock Paris-lille-3d: a large and high-quality ground truth urban point cloud dataset for automatic segmentation and classification.
\newblock \emph{CoRR}, abs/1712.00032, 2017.

\bibitem[Tan et~al.(2020)Tan, Qin, Ma, Li, Du, Cai, Yang, and Li]{tan2020toronto3d}
Weikai Tan, Nannan Qin, Lingfei Ma, Ying Li, Jing Du, Guorong Cai, Ke Yang, and Jonathan Li.
\newblock Toronto-3d: A large-scale mobile lidar dataset for semantic segmentation of urban roadways.
\newblock \emph{CVPR Workshops}, pages 202--203, 2020.

\bibitem[Thomas et~al.(2019)Thomas, Qi, Deschaud, Marcotegui, Goulette, and Guibas]{thomas2019kpconv}
Hugues Thomas, Charles~R Qi, Jean-Emmanuel Deschaud, Beatriz Marcotegui, Fran{\c{c}}ois Goulette, and Leonidas~J Guibas.
\newblock Kpconv: Flexible and deformable convolution for point clouds.
\newblock In \emph{Proceedings of the IEEE/CVF International Conference on Computer Vision (ICCV)}, pages 6411--6420, 2019.

\bibitem[Tylecek and {\v{S}}{\'a}ra(2013)]{tylecek2013spatial}
Radim Tylecek and Radim {\v{S}}{\'a}ra.
\newblock Spatial pattern templates for recognition of objects with regular structure.
\newblock In \emph{Proc. German Conference on Pattern Recognition (GCPR)}, Saarbr{\"u}cken, Germany, 2013.

\bibitem[Vu et~al.(2022)Vu, Kim, Luu, Nguyen, and Yoo]{vu2022softgroup}
Thang Vu, Kookhoi Kim, Tung~M. Luu, Xuan~Thanh Nguyen, and Chang~D. Yoo.
\newblock {SoftGroup} for 3d instance segmentation on point clouds.
\newblock In \emph{Proceedings of the IEEE/CVF Conference on Computer Vision and Pattern Recognition (CVPR)}, pages 2708--2717, 2022.

\bibitem[Wang et~al.(2019)Wang, Sun, Liu, Sarma, Bronstein, and Solomon]{wang2019dynamic}
Yue Wang, Yongbin Sun, Ziwei Liu, Sanjay~E Sarma, Michael~M Bronstein, and Justin~M Solomon.
\newblock Dynamic graph cnn for learning on point clouds.
\newblock \emph{ACM Transactions on Graphics}, 38\penalty0 (5):\penalty0 1--12, 2019.

\bibitem[Wang et~al.(2026)Wang, Wang, Yang, Gyawali, Xu, Klimkowska, Jing, Biljecki, Holst, Busam, Sheil, and Wysocki]{wang2026underonefacade}
Yi Wang, Fan Wang, Wanru Yang, Prabin Gyawali, Ziyang Xu, Anna Klimkowska, Yixiong Jing, Filip Biljecki, Christoph Holst, Benjamin Busam, Brian Sheil, and Olaf Wysocki.
\newblock Underonefacade: Worldwide facade semantic segmentation benchmark dataset, 2026.
\newblock ECCV 2026.

\bibitem[Wu et~al.(2022)Wu, Lao, Jiang, Liu, and Zhao]{wu2022ptv2}
Xiaoyang Wu, Yixing Lao, Li Jiang, Xihui Liu, and Hengshuang Zhao.
\newblock Point transformer v2: Grouped vector attention and partition-based pooling.
\newblock In \emph{Advances in Neural Information Processing Systems (NeurIPS)}, 2022.

\bibitem[Wu et~al.(2024)Wu, Jiang, Wang, Liu, Liu, Qiao, Ouyang, He, and Zhao]{wu2024ptv3}
Xiaoyang Wu, Li Jiang, Peng-Shuai Wang, Zhijian Liu, Xihui Liu, Yu Qiao, Wanli Ouyang, Tong He, and Hengshuang Zhao.
\newblock Point transformer v3: Simpler, faster, stronger.
\newblock In \emph{Proceedings of the IEEE/CVF Conference on Computer Vision and Pattern Recognition (CVPR)}, 2024.

\bibitem[Wysocki et~al.(2022)Wysocki, Hoegner, and Stilla]{wysockiTUMFACADE}
O. Wysocki, L. Hoegner, and U. Stilla.
\newblock Tum-faÇade: Reviewing and enriching point cloud benchmarks for faÇade segmentation.
\newblock \emph{The International Archives of the Photogrammetry, Remote Sensing and Spatial Information Sciences}, XLVI-2/W1-2022:\penalty0 529--536, 2022.

\bibitem[Wysocki et~al.(2025)Wysocki, Tan, Froech, Xia, Wysocki, Hoegner, Cremers, and Holst]{wysocki2024zahaintroducinglevelfacade}
Olaf Wysocki, Yue Tan, Thomas Froech, Yan Xia, Magdalena Wysocki, Ludwig Hoegner, Daniel Cremers, and Christoph Holst.
\newblock Zaha: Introducing the level of facade generalization and the large-scale point cloud facade semantic segmentation benchmark dataset.
\newblock In \emph{Proceedings of the IEEE/CVF Winter Conference on Applications of Computer Vision (WACV)}, 2025.

\bibitem[Yang et~al.(2023)Yang, Xue, Zhang, Xie, Fu, and Huang]{yang2023urbannis}
Guoqing Yang, Fuyou Xue, Qi Zhang, Ke Xie, Chi-Wing Fu, and Hui Huang.
\newblock {UrbanBIS}: a large-scale benchmark for fine-grained urban building instance segmentation.
\newblock \emph{ACM Transactions on Graphics}, 42\penalty0 (4), 2023.

\bibitem[Zhao et~al.(2021)Zhao, Jiang, Jia, Torr, and Koltun]{zhao2021point}
Hengshuang Zhao, Li Jiang, Jiaya Jia, Philip~HS Torr, and Vladlen Koltun.
\newblock Point transformer.
\newblock In \emph{ICCV}, pages 16259--16268, 2021.

\bibitem[Zhu et~al.(2020)Zhu, Gehrung, Huang, Borgmann, Sun, Hoegner, Hebel, Xu, and Stilla]{zhu2020tum}
Jingwei Zhu, Joachim Gehrung, Rong Huang, Bj{\"o}rn Borgmann, Zhenghao Sun, Ludwig Hoegner, Marcus Hebel, Yusheng Xu, and Uwe Stilla.
\newblock Tum-mls-2016: An annotated mobile lidar dataset of the tum city campus for semantic point cloud interpretation in urban areas.
\newblock \emph{Remote Sensing}, 12\penalty0 (11):\penalty0 1875, 2020.

\end{thebibliography}
   }
   \clearpage
   \appendix
   \clearpage
\maketitlesupplementary

\section{Additional Dataset Details}
\label{appendix:dataset}

This section provides additional details on HFX3D beyond those included in the main paper,
including capture diversity, annotated facade examples, annotation review statistics,
building-level composition, and the train/validation/test split composition.

\subsection{Environmental Diversity}
\label{appendix:environment}

\begin{figure}[t]
\centering
\setlength{\tabcolsep}{1pt}
\begin{tabular}{ccc}
  \includegraphics[width=0.32\linewidth]{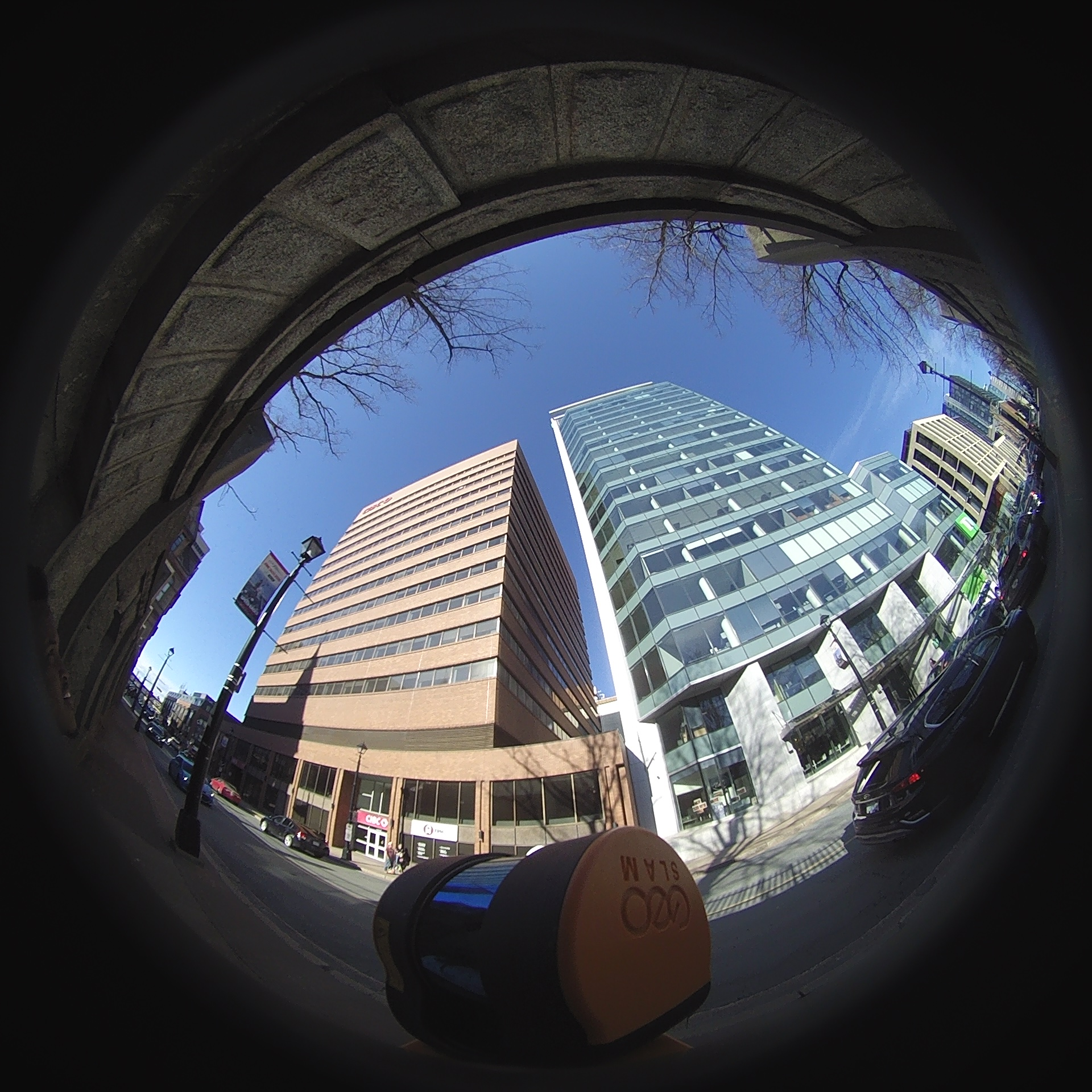} &
  \includegraphics[width=0.32\linewidth]{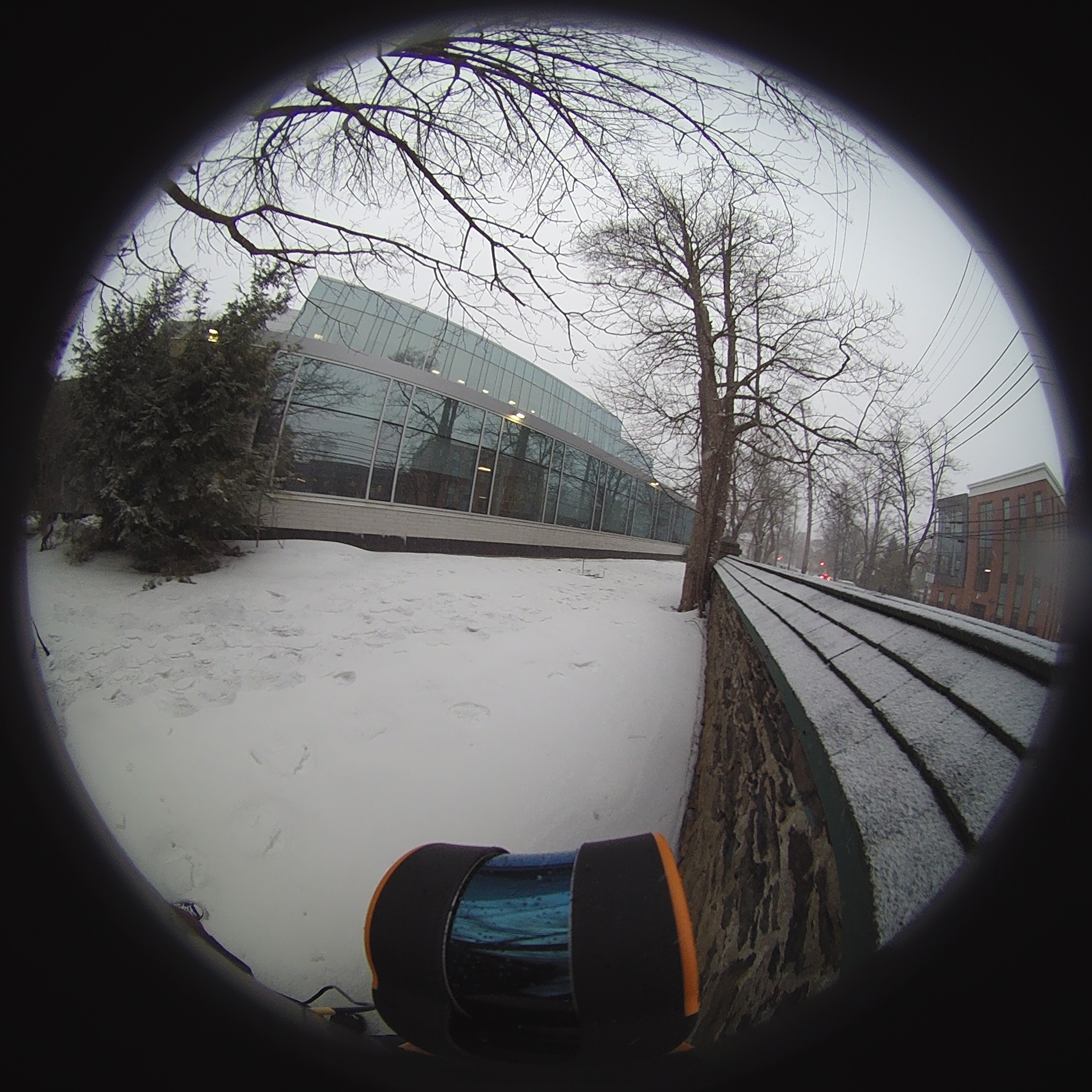} &
  \includegraphics[width=0.32\linewidth]{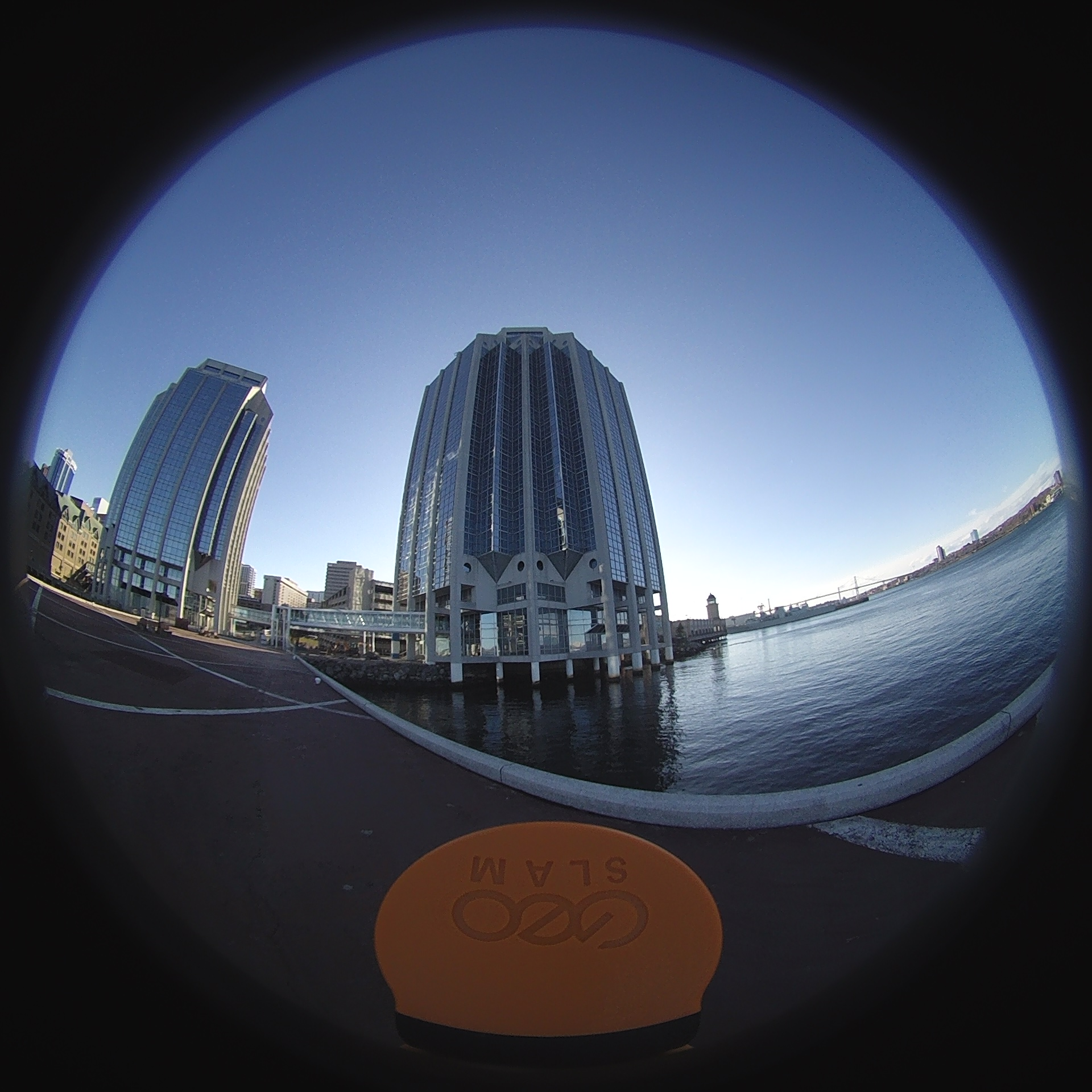} \\
  \includegraphics[width=0.32\linewidth]{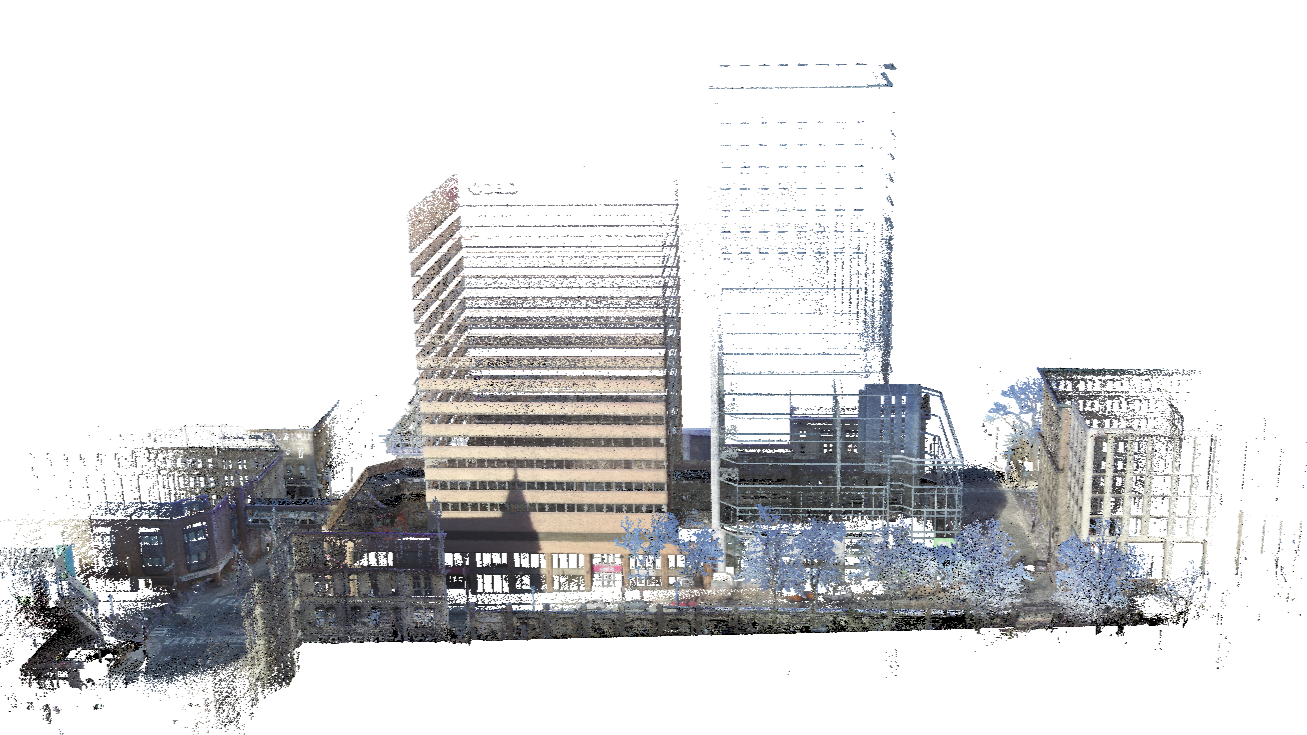} &
  \includegraphics[width=0.32\linewidth]{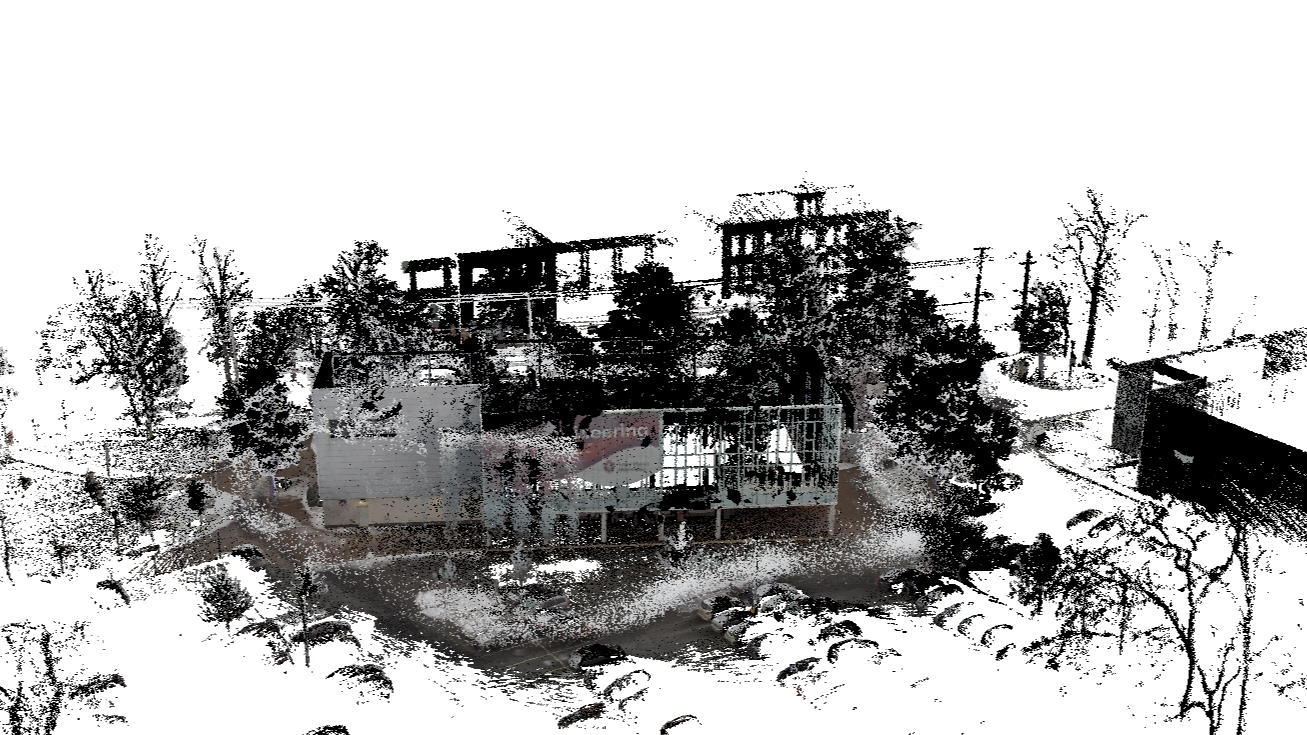} &
  \includegraphics[width=0.32\linewidth]{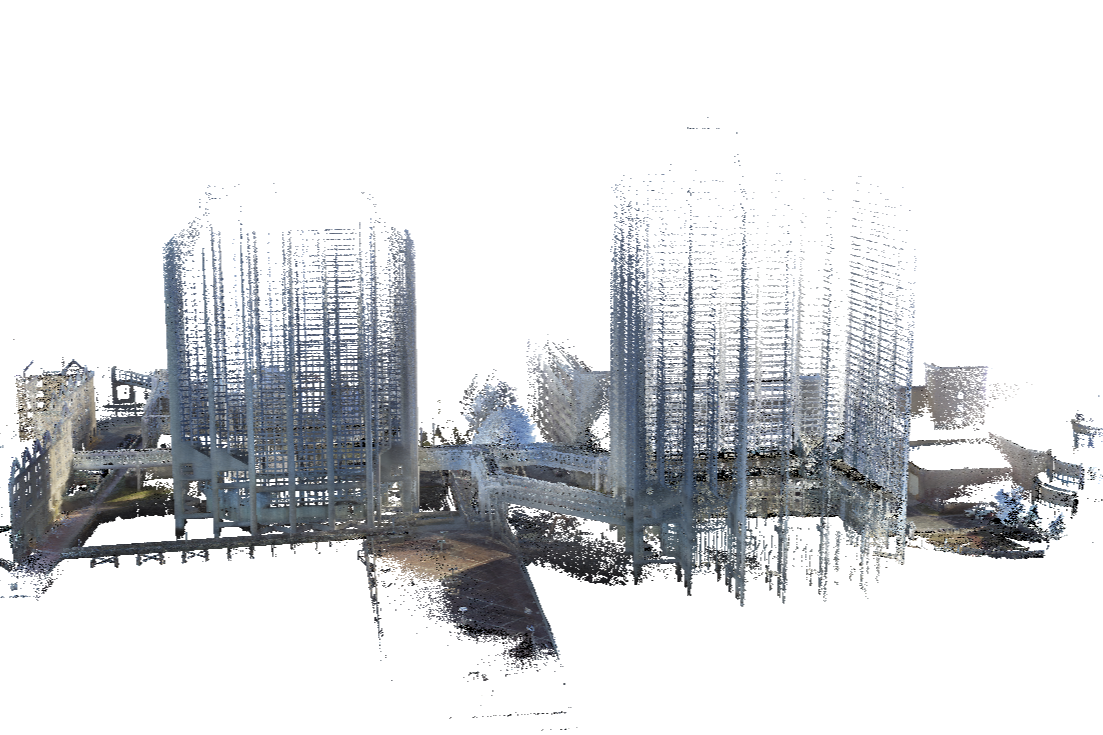} \\
\end{tabular}
\caption{\textbf{ZEB Vision fisheye frames and corresponding RGB-coloured LiDAR point clouds.}
Top row: fisheye frames for fall, winter, and waterfront conditions.
Bottom row: corresponding RGB-coloured point clouds from camera-LiDAR fusion.}
\label{fig:fisheye_pairs}
\end{figure}

\begin{figure}[t]
\centering
\setlength{\tabcolsep}{2pt}
\begin{tabular}{cc}
  \includegraphics[width=0.48\columnwidth]{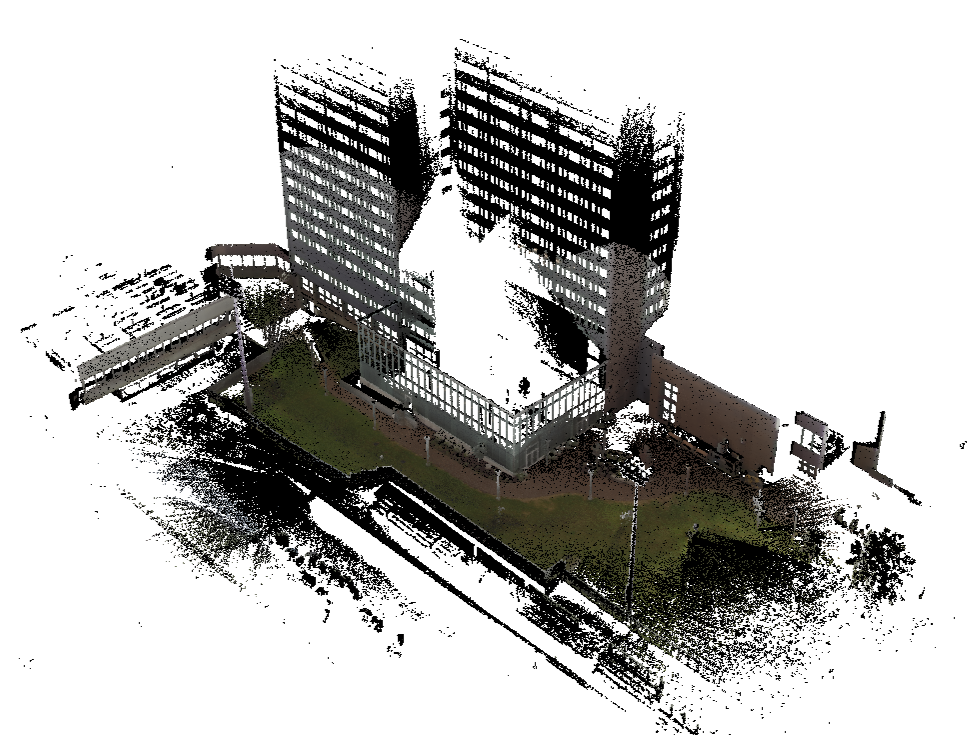} &
  \includegraphics[width=0.48\columnwidth]{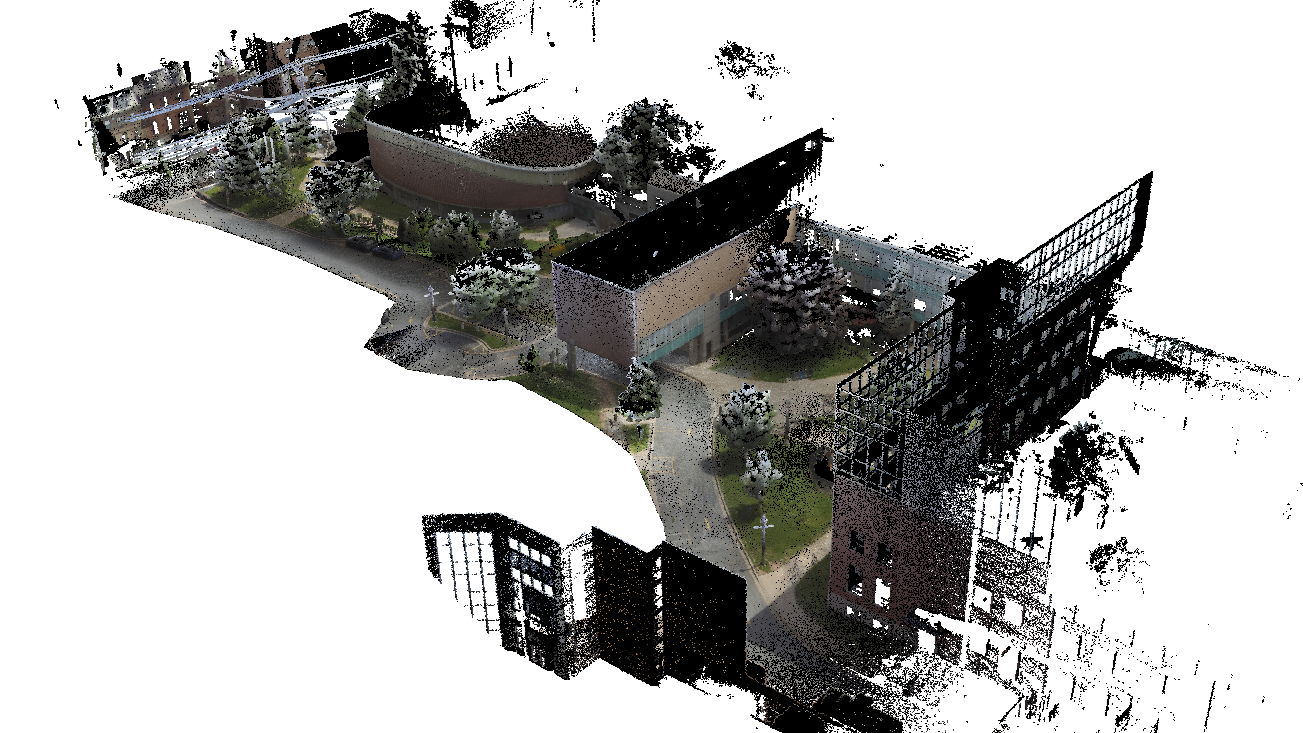} \\
  \footnotesize (a) Spring &
  \footnotesize (b) Summer \\[4pt]

  \includegraphics[width=0.48\columnwidth]{figures/cond_fall_front_face.png} &
  \includegraphics[width=0.48\columnwidth]{figures/cond_winter_front_face.png} \\
  \footnotesize (c) Fall &
  \footnotesize (d) Winter \\[4pt]

  \includegraphics[width=0.48\columnwidth]{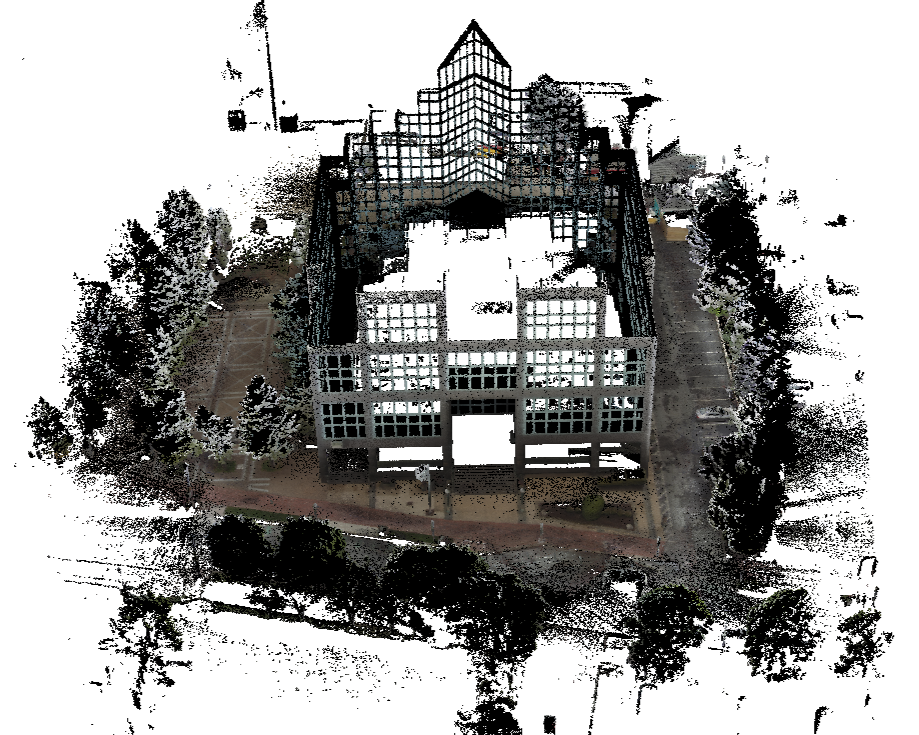} &
  \includegraphics[width=0.48\columnwidth]{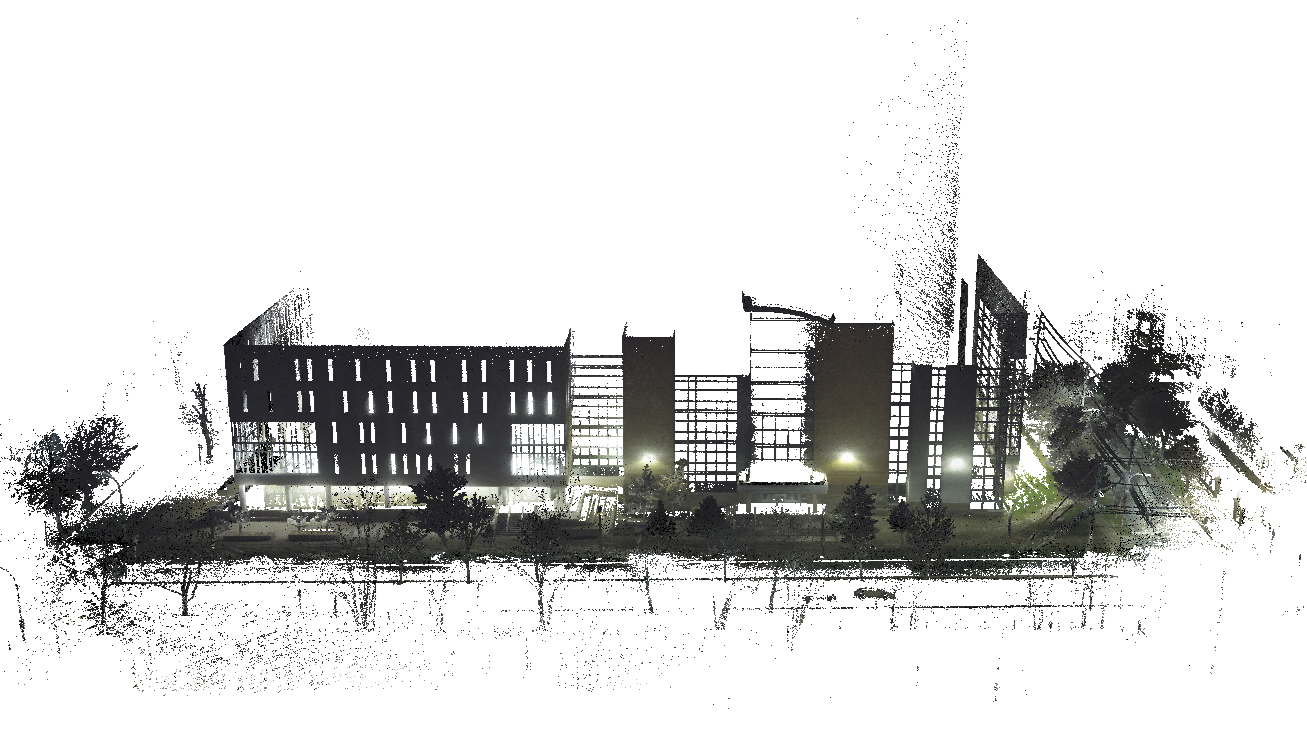} \\
  \footnotesize (e) Daytime &
  \footnotesize (f) Night-time \\
\end{tabular}

\caption{\textbf{Environmental diversity of HFX3D.}
RGB-coloured LiDAR scans of the same facades across conditions: (a)~spring, (b)~summer,
(c)~fall, (d)~winter, (e)~daytime, and (f)~night-time. Seasonal change is visible in vegetation
density, fall foliage, and snow cover; night-time scans retain full geometric detail, with
colour driven by facade and street lighting.}
\label{fig:conditions}
\end{figure}

HFX3D was captured over six to eight months, spanning all four seasons, both daytime and
night-time conditions, and varied weather including snow, fog, and clear sun. To our
knowledge, this environmental range is absent from existing facade benchmarks, which are
typically captured under a single, narrow set of conditions. Figure~\ref{fig:fisheye_pairs}
shows raw ZEB Vision fisheye frames with their corresponding RGB-coloured point clouds, while
Figure~\ref{fig:conditions} shows the same facades captured across seasons and lighting
conditions.

\subsection{Annotation Review and Dataset Statistics}
\label{appendix:review_stats}

\noindent\textbf{Annotation review and instance refinement.}
Table~\ref{tab:review_stats} reports, for each of the 26 buildings, Cohen's Kappa between the original and reviewed semantic labels after the second cross-review round, the DBSCAN neighbourhood radius $\epsilon$ used for instance clustering, and the candidate instance count before and after manual refinement in CloudCompare. After first-round corrections were applied and re-reviewed, agreement is near-perfect across the dataset, with a mean $\kappa$ of 0.99 and no building below 0.81. Refinement predominantly merged over-segmented clusters.

\begin{table*}[!t]
\centering
\caption{\textbf{Per-building annotation review and instance clustering statistics.} $\kappa$: Cohen's Kappa between original and reviewed semantic labels after the second cross-review round. $\epsilon$: DBSCAN neighbourhood radius used for instance clustering. Instances: candidate count from automatic clustering $\to$ final count after manual refinement.}
\label{tab:review_stats}
\small
\begin{tabular}{lccc c lccc}
\toprule
Building & $\kappa$ & $\epsilon$ & Instances & \phantom{x} & Building & $\kappa$ & $\epsilon$ & Instances \\
\midrule
BLD001 & 1.00 & 0.15 & 273$\to$264 & & BLD014 & 1.00 & 0.27 & 207$\to$170 \\
BLD002 & 1.00 & 0.50 & 352$\to$310 & & BLD015 & 1.00 & 0.70 & 79$\to$78 \\
BLD003 & 0.81 & 0.33 & 57$\to$53 & & BLD016 & 1.00 & 0.34 & 116$\to$60 \\
BLD004 & 1.00 & 0.33 & 140$\to$138 & & BLD017 & 1.00 & 0.17 & 435$\to$350 \\
BLD005 & 1.00 & 0.30 & 445$\to$375 & & BLD018 & 1.00 & 0.30 & 167$\to$180 \\
BLD006 & 1.00 & 0.16 & 467$\to$416 & & BLD019 & 1.00 & 0.30 & 156$\to$152 \\
BLD007 & 1.00 & 0.33 & 96$\to$79 & & BLD020 & 1.00 & 0.09 & 1139$\to$436 \\
BLD008 & 1.00 & 0.20 & 165$\to$149 & & BLD021 & 0.98 & 0.50 & 159$\to$94 \\
BLD009 & 1.00 & 0.27 & 258$\to$194 & & BLD022 & 1.00 & 0.35 & 192$\to$175 \\
BLD010 & 0.99 & 0.13 & 881$\to$483 & & BLD023 & 1.00 & 0.33 & 196$\to$176 \\
BLD011 & 1.00 & 0.20 & 990$\to$873 & & BLD024 & 1.00 & 0.30 & 333$\to$218 \\
BLD012 & 1.00 & 0.30 & 369$\to$355 & & BLD025 & 1.00 & 0.33 & 89$\to$74 \\
BLD013 & 1.00 & 0.33 & 330$\to$139 & & BLD026 & 0.91 & 0.40 & 495$\to$427 \\
\bottomrule
\end{tabular}
\end{table*}

\begin{figure}[t]
\centering
\includegraphics[width=\linewidth]{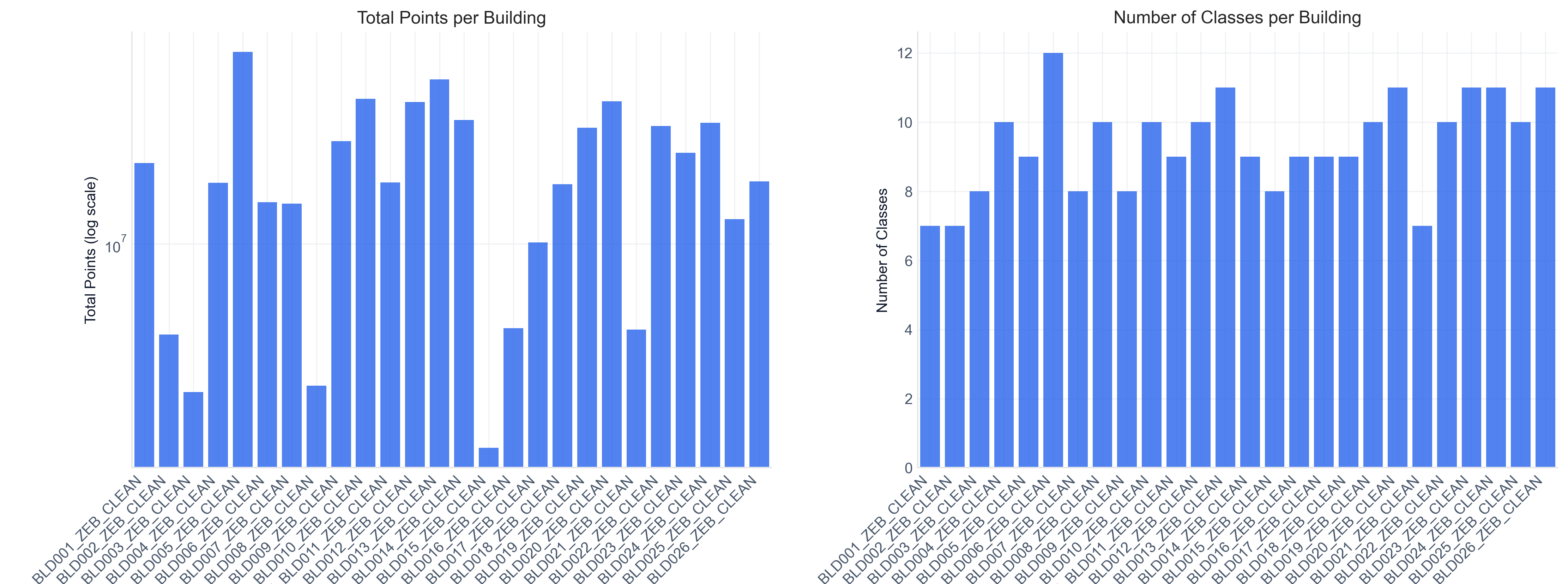}
\caption{\textbf{Building-level statistics.}
\emph{Left:} point count per building (log scale).
\emph{Right:} semantic classes present per building, reflecting architectural diversity across the dataset.}
\label{fig:file_stats}
\end{figure}

\begin{figure}[t]
\centering
\includegraphics[width=\linewidth]{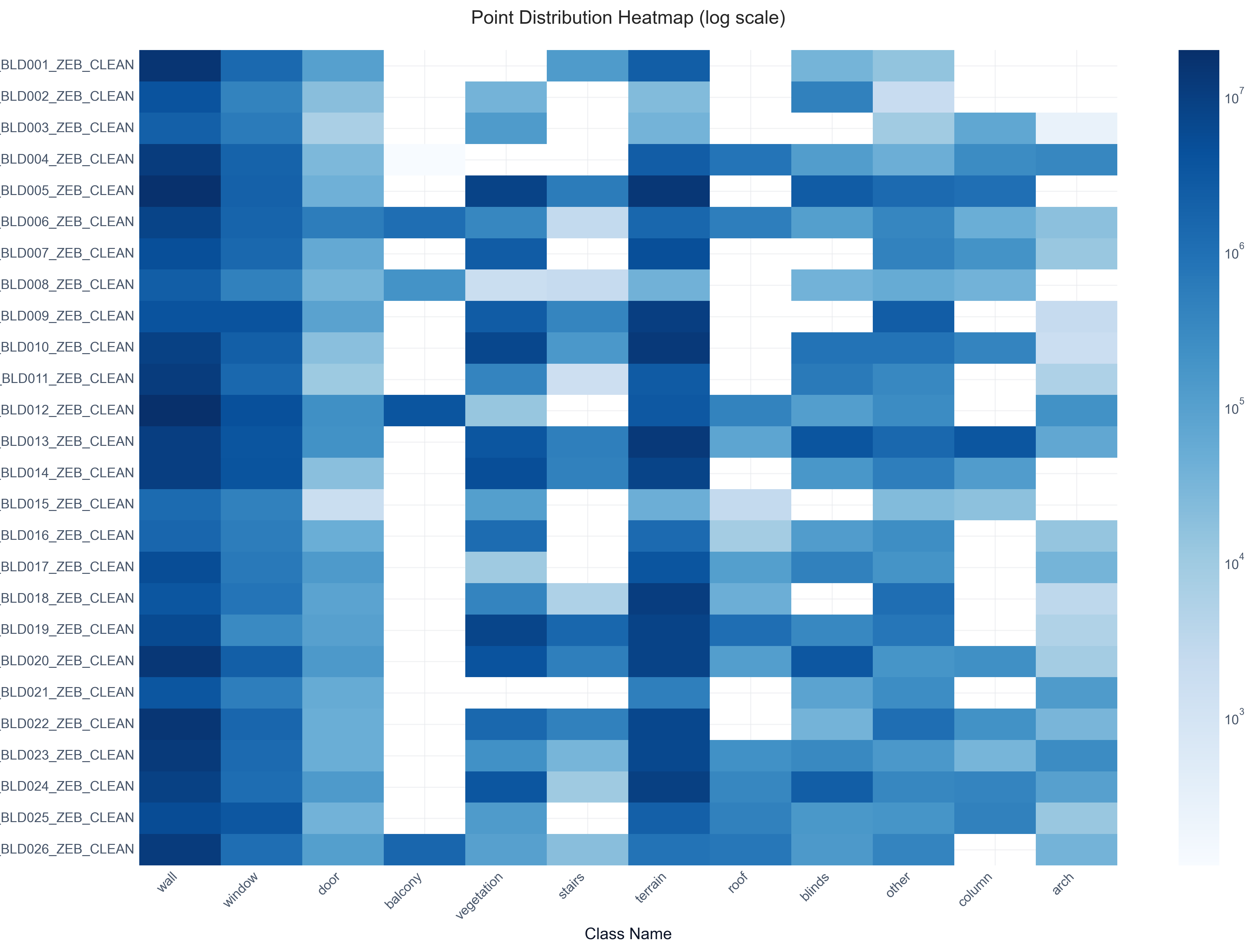}
\caption{\textbf{Per-building class presence heatmap (log scale).}
Rows: 26 buildings; columns: 12 semantic classes. White cells indicate a class is absent.
Rare classes such as balcony, arch, and stairs are missing from several buildings,
compounding long-tail imbalance and motivating careful split design.}
\label{fig:heatmap}
\end{figure}

\noindent\textbf{Building-level composition.}
HFX3D exhibits substantial variation in both scene size and semantic composition.
Figure~\ref{fig:file_stats} shows that individual buildings range from fewer than
2M to more than 38M points and contain between 7 and 12 semantic classes.
Figure~\ref{fig:heatmap} shows the building-level distribution of classes.
Rare categories such as balcony, arch, and stairs occur in only a subset of buildings,
producing a pronounced long-tail distribution.

\begin{figure}[t]
\centering
\includegraphics[width=\linewidth]{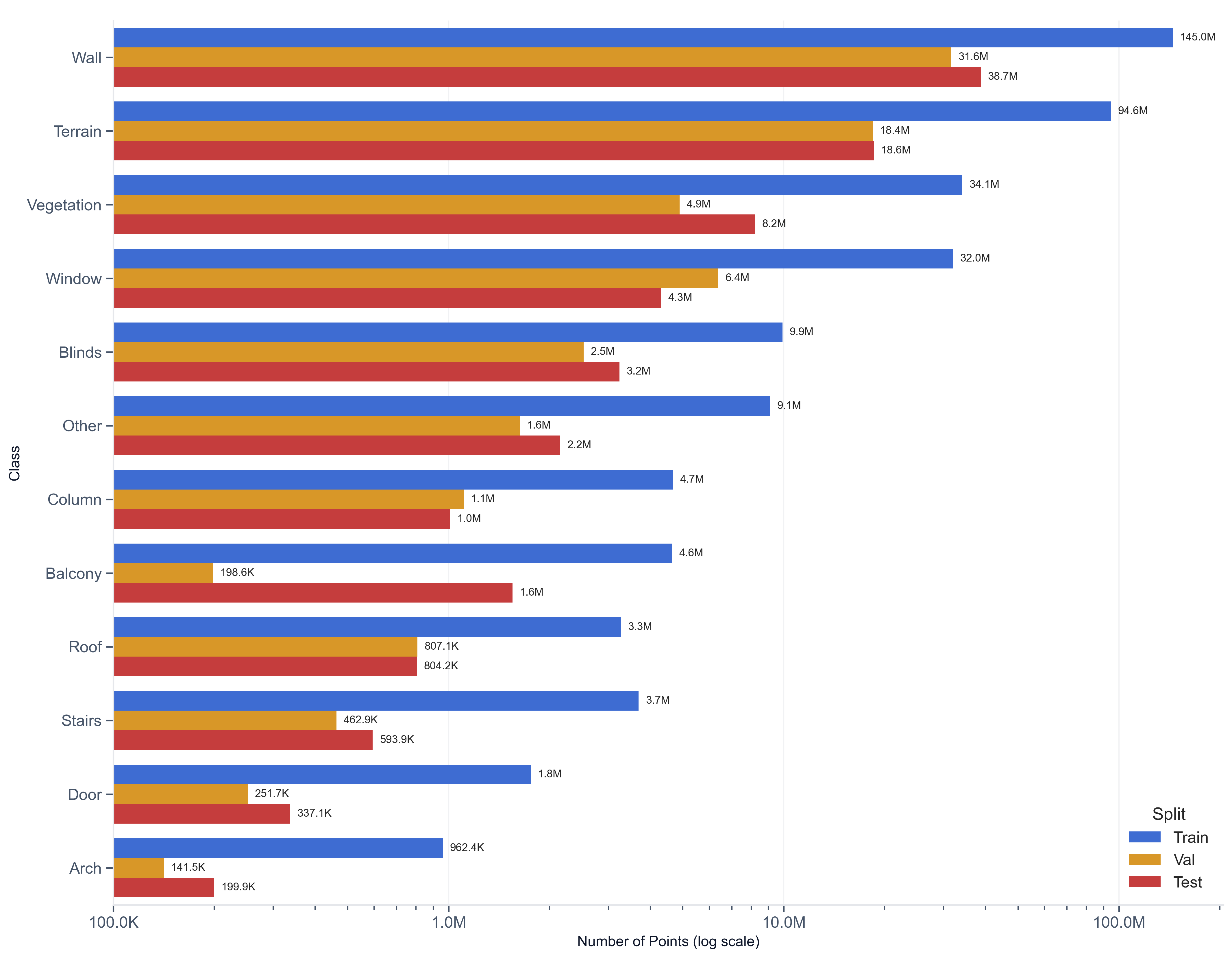}
\caption{\textbf{Per-class point counts across train / val / test splits (log scale).}
Wall and terrain are well-represented across all splits.
Balcony, arch, and stairs are severely underrepresented in validation.}
\label{fig:splits}
\end{figure}

\noindent\textbf{Train/validation/test composition.}
Figure~\ref{fig:splits} reports the per-class point distribution across the
train, validation, and test splits. Wall and terrain are well represented throughout,
whereas balcony, arch, and stairs remain sparse, particularly in validation.
This imbalance is a direct consequence of the building-level partition: because entire
buildings rather than individual points are assigned to splits, classes occurring in
only a few buildings cannot be uniformly distributed without violating spatial
independence between training and evaluation scenes.

\subsection{Instance Taxonomy and Counts}
\label{appendix:instance_taxonomy}

Instance AP is only meaningful for classes whose objects can be counted. Following the
ScanNet~\cite{dai2017scannet} convention, we therefore split the 12 classes into countable
``thing'' classes and continuous ``stuff'' classes, and score only the thing classes with
instance AP. The eight thing classes are window, door, balcony, column, arch, blinds,
vegetation, and stairs; the four stuff classes are wall, terrain, roof, and other.

Stuff classes are excluded from instance AP because they have no consistent notion of a single
object. A facade may reasonably be annotated as one wall or as many, so an instance score for
wall would reflect this annotation choice rather than model quality. Stuff classes still retain
instance identifiers in the released data and are scored semantically. Two thing classes need a
counting convention: vegetation is one instance per bush or tree, and a staircase is a single
instance rather than one per flight.

For scoring, ground-truth thing instances smaller than 200 points are marked ignore and treated
as void, and predicted instances smaller than 100 points are discarded, following ScanNet's
minimum region size. Table~\ref{tab:thing_counts} lists the resulting 5{,}268 scored thing
instances per class and split.

\begin{table}[t]
\centering
\caption{\textbf{Thing-class instance counts} across all 26 buildings, per split, after the
200-point minimum-instance filter. The four stuff classes remain
semantically labeled but are excluded from instance AP.}
\label{tab:thing_counts}
\begin{tabular}{lrrrr}
\toprule
Class & Train & Val & Test & Total \\
\midrule
Window     & 1{,}835 & 175 & 638 & 2{,}648 \\
Blinds     & 1{,}195 & 120 & 182 & 1{,}497 \\
Vegetation &     249 &  25 &  63 &     337 \\
Balcony    &     177 &  21 &  53 &     251 \\
Door       &     119 &  15 &  64 &     198 \\
Arch       &     118 &  28 &  22 &     168 \\
Column     &      78 &  37 &  14 &     129 \\
Stairs     &      29 &   6 &   5 &      40 \\
\midrule
Total      & 3{,}800 & 427 & 1{,}041 & 5{,}268 \\
\bottomrule
\end{tabular}
\end{table}

\subsection{Annotation Examples}
\label{appendix:examples}

\begin{figure}[t]
\centering
\includegraphics[width=\linewidth]{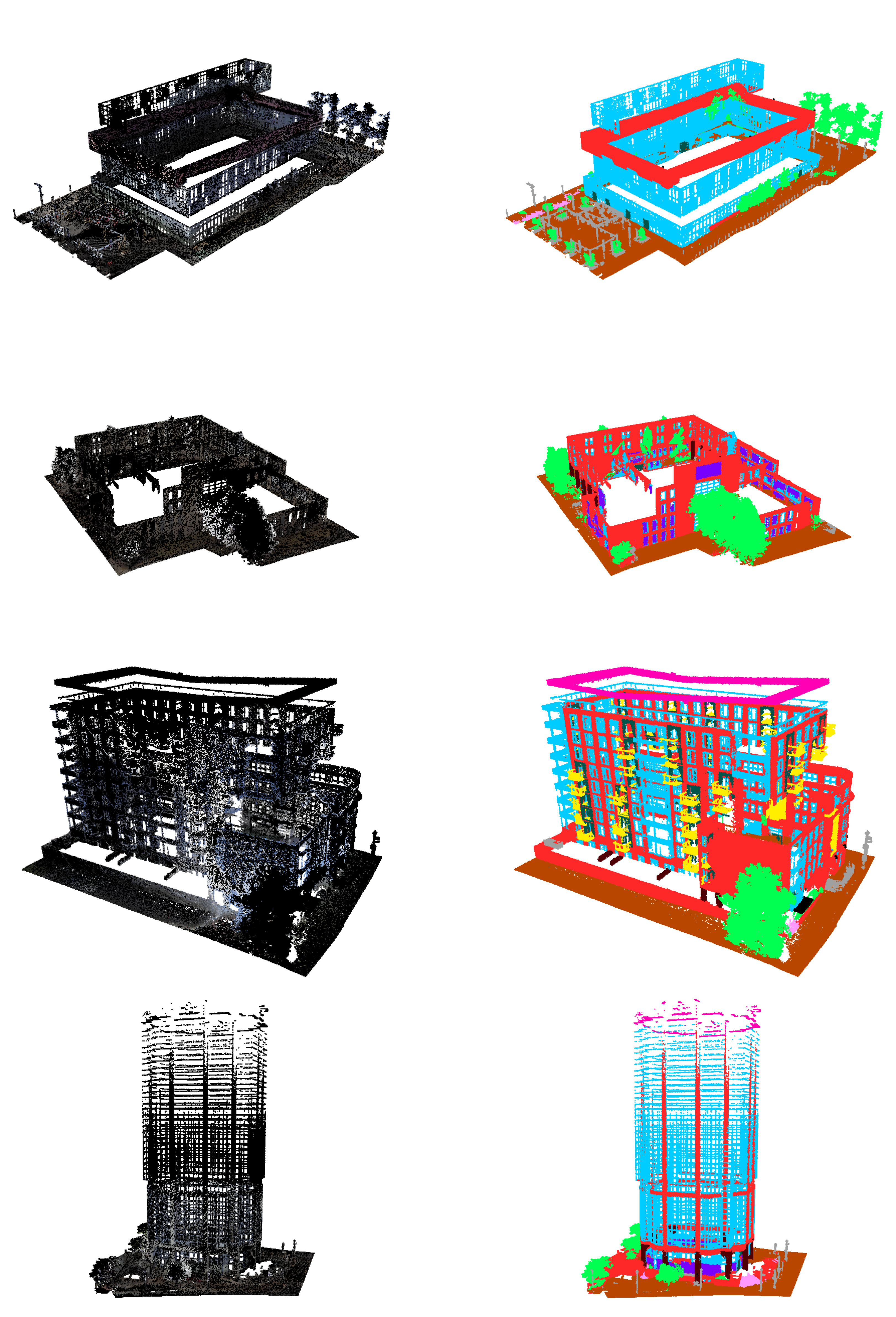}
\caption{Additional annotated facade examples from HFX3D.
Rows correspond to (top to bottom): a commercial glass-curtain
building, a heritage masonry building, a mid-rise residential
block, and a government institutional building. Each panel
shows the raw colored point cloud (left) and the
semantically annotated version (right).}
\label{fig:appendix_facades}
\end{figure}

Figure~\ref{fig:appendix_facades} shows four additional annotated facades spanning distinct architectural styles that differ in material, glazing ratio, and structural geometry.

\section{Functional Attribute Priors and Rules}
\label{appendix:func_rules}

Each semantic class maps to a fixed set of candidate functional attributes
(the semantic-to-functional candidate mapping, Fig.~4 of the main paper), and
each candidate carries a semantic prior $P(a\mid c)$. These priors are expert
heuristics set from architectural knowledge and fixed before the human review
round; attributes outside a class's candidate set are fixed to $0$.
Table~\ref{tab:priors} lists the priors for every class. Instance-dependent
attributes are then refined from geometric evidence $G$ and contextual evidence
$C$, fused with the prior in the odds domain as described in the main paper;
Table~\ref{tab:evidence} specifies the evidence used for each refined attribute.

\begin{table}[t]
\centering
\caption{\textbf{Semantic priors $P(a\mid c)$ per class.} Attributes not listed for a
class are fixed to $0$. Abbreviations as in Fig.~4 of the main paper.}
\label{tab:priors}
\small
\begin{tabular}{@{}ll@{}}
\toprule
\textbf{Class} & \textbf{Candidate attributes (prior)} \\
\midrule
Wall       & LB .65, TE .90, VS .40 \\
Roof       & LB .85, TE .90, DR .40, FP .75 \\
Column     & LB .90, AE .50 \\
Arch       & LB .35, AE .25, IL .35, SU .45 \\
Window     & TE .90, OP .85, SS .70, VE .75, NL .90 \\
Door       & AC .90, OP .90, TE .90 \\
Blinds     & SS .90, OP .90, PS .90 \\
Balcony    & LB .90, AC .90, VS .10, FP .90 \\
Stairs     & CI .90, FP .90 \\
Terrain    & CI .85 \\
Vegetation & AE .90 \\
Other      & \textit{none} \\
\bottomrule
\end{tabular}
\end{table}

\begin{table*}[t]
\centering
\footnotesize
\caption{\textbf{Evidence used to refine instance-dependent attributes.}
$G$ = instance geometry, $C$ = context. Pairs not listed are decided by the prior
alone ($G=C=0.5$).}
\label{tab:evidence}
\begin{tabular}{@{}l p{0.38\textwidth} p{0.38\textwidth}@{}}
\toprule
\textbf{Class $\cdot$ attribute} & \textbf{$G$ (geometry)} & \textbf{$C$ (context)} \\
\midrule
Wall $\cdot$ LB          & ground continuity & support from above \\
Wall, Balcony $\cdot$ VS & -- & vegetation alongside or on top \\
Window $\cdot$ OP        & depth/normal variability or handle-like protrusion & -- \\
Window $\cdot$ SS        & -- & adjacent blinds \\
Window $\cdot$ VE        & operability score & -- \\
Door $\cdot$ TE          & position on building outline & host wall \\
Balcony $\cdot$ LB       & -- & wall attachment or support from below \\
Stairs $\cdot$ FP        & elevation $\times$ exposed edge & barrier \\
Terrain $\cdot$ CI       & low slope & adjacent door or stairs \\
Roof $\cdot$ LB          & -- & support from below \\
Roof $\cdot$ DR          & slope $\times$ edge & downhill neighbour \\
Roof $\cdot$ FP          & low slope $\times$ exposed edge & barrier \\
Blinds $\cdot$ OP        & depth/normal variability & adjacent window \\
Blinds $\cdot$ PS        & -- & window coverage \\
Column $\cdot$ LB        & verticality $\times$ ground continuity & support from above \\
Column $\cdot$ AE        & enlarged top (capital) & -- \\
Arch $\cdot$ LB          & opening shape & host wall and support \\
Arch $\cdot$ AE          & shallow closed shape & host wall \\
Arch $\cdot$ IL, SU      & compact fixture-like cluster & host wall / elevation \\
\bottomrule
\end{tabular}
\end{table*}

Context relations are computed on a graph that links instances whose bounding
boxes lie within $1.5$~m or that are among each other's $12$ nearest centroids;
all constants are provided in the released configuration. Reviewers were not
restricted to these candidate sets: they could assign any attribute to an instance
even if the pipeline had not proposed it.

\section{Implementation Details}
\label{appendix:impl}

\noindent\textbf{Semantic segmentation.} All seven semantic architectures train for 200 epochs, selecting the best-validation checkpoint. Training runs on the Tamia cluster's NVIDIA H100 GPUs inside Apptainer containers (PyTorch 2.4.1, CUDA 12.1). Each model keeps its own author-recommended recipe rather than a forced common one; the shared 7-channel input and 100\%-coverage, augmentation-free evaluation follow the main paper.

\noindent\textbf{Instance segmentation.} The three learned models train from scratch for 18{,}000 steps at batch 4 on the same Tamia H100 GPUs, with test evaluation on the Vulcan cluster's L40S GPUs. Owing to compute constraints we report a single run per method, with the checkpoint selected on validation near the end of training, so scores may still be improving. ISBNet uses width 16 and SoftGroup width 32, following their published recipes. Full buildings exceed memory (up to roughly 48M points), so inference runs on overlapping vertical-column crops roughly 15~m wide with about 5~m overlap, each subsampled to at most 250k points. Predicted masks are then propagated to all tile points and merged across tiles by connected-component IoU-union. Scoring uses a ScanNet-v2-convention~\cite{dai2017scannet} evaluator: AP averaged over IoU 0.50 to 0.90 in steps of 0.05, plus AP50 and AP25.

\noindent\textbf{Clustering baselines.} All four clustering methods run per class on 5~cm voxels with the same evaluator; standardisation and one validation-selected scale parameter per method come from training-split statistics. Because they output no per-instance confidence scores, their AP is measured at a single precision--recall operating point rather than over a ranked prediction list.

\section{Additional Qualitative Analysis} \label{appendix:qualitative}

\begin{figure*}[t]
\centering
\includegraphics[width=\linewidth]{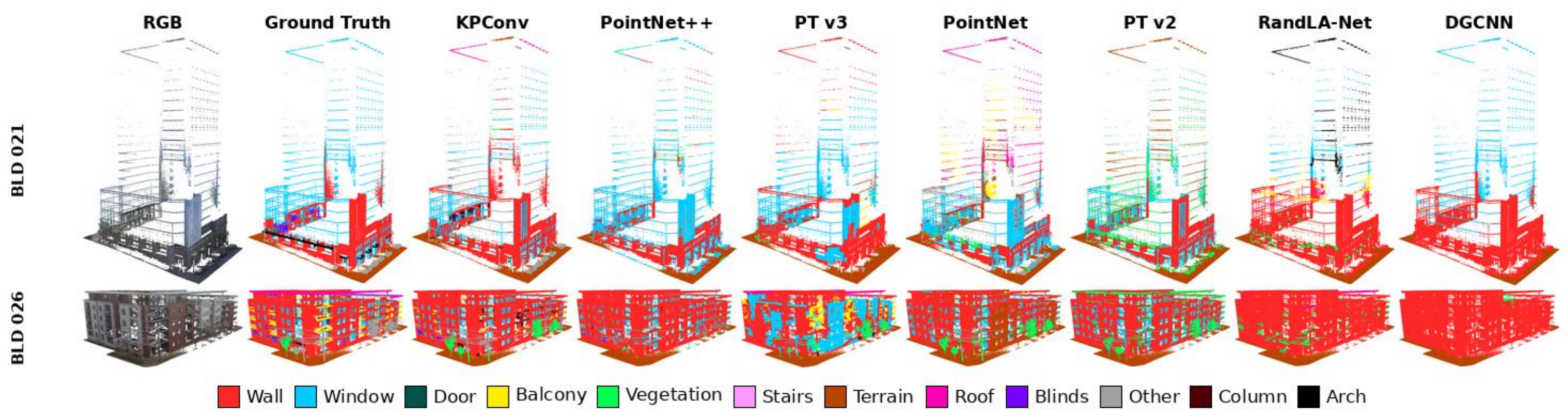}
\caption{\textbf{Qualitative semantic segmentation on two test buildings.}
Rows: two held-out test buildings. Columns: the RGB scan, ground truth, and predictions
from all seven backbones ordered left-to-right by decreasing mIoU (KPConv to DGCNN).}
\label{fig:qual_semantic_grid}
\end{figure*}

Figure~\ref{fig:qual_semantic_grid} shows qualitative semantic segmentation predictions on
two test buildings for all seven benchmarked backbones. The ranking is visible throughout:
predictions retain progressively less fine structure from KPConv to DGCNN, with the weakest
models reducing each facade to its dominant wall and terrain classes.

\begin{figure*}[t]
\centering
\includegraphics[width=\linewidth]{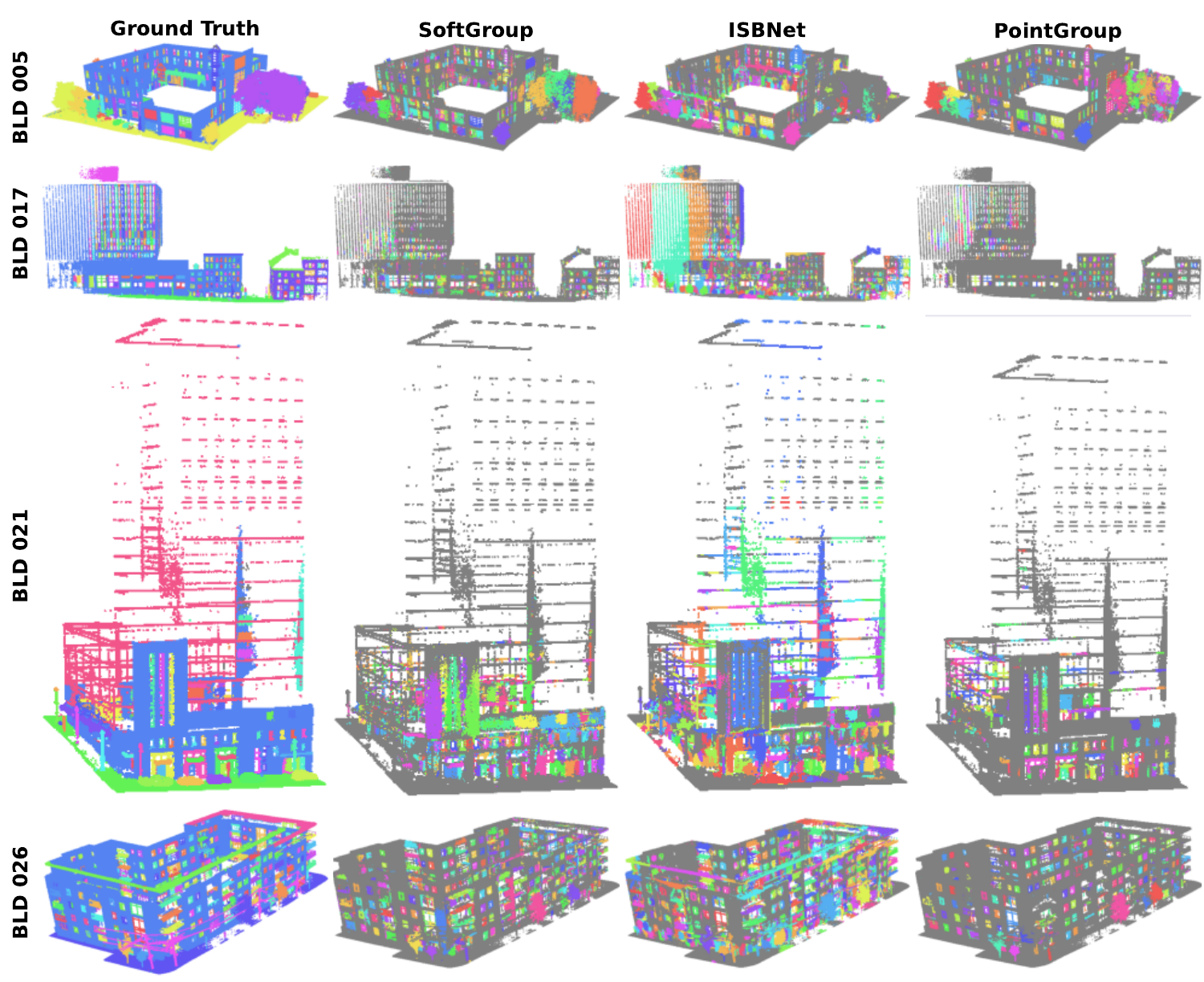}
\caption{\textbf{Qualitative instance segmentation, learned methods.}
Rows: four test buildings. Columns: ground truth followed by SoftGroup, ISBNet, and
PointGroup predictions (best to worst by AP). All three methods, trained end-to-end without
semantic labels, struggle to delineate repetitive facade elements, with SoftGroup the
strongest (Table~4).}
\label{fig:qual_instance_learned}
\end{figure*}

\begin{figure*}[t]
\centering
\includegraphics[width=\linewidth]{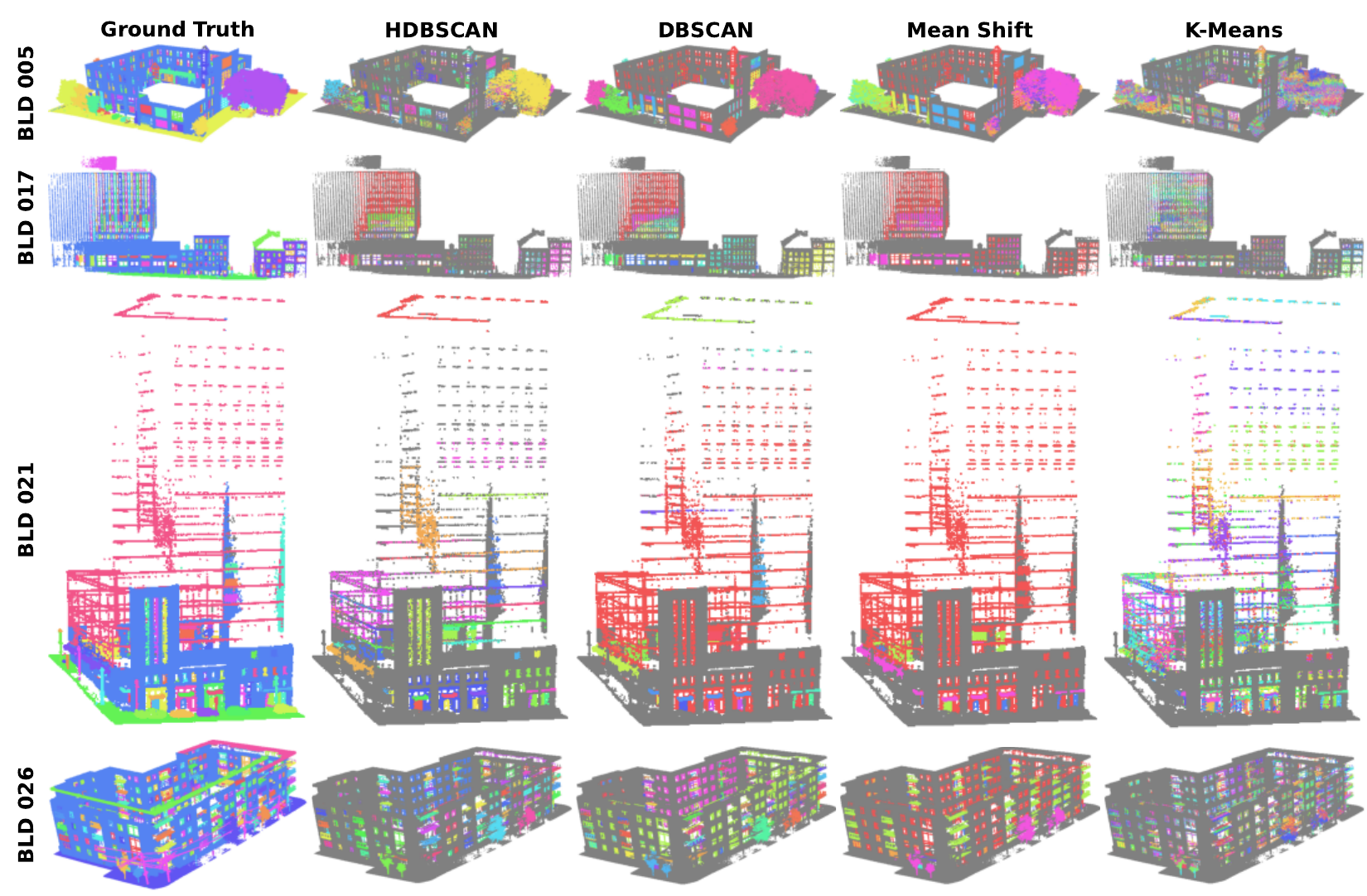}
\caption{\textbf{Qualitative instance segmentation, geometry-based clustering given
ground-truth semantics.} Rows: four test buildings. Columns: ground truth followed by
HDBSCAN, DBSCAN, Mean Shift, and K-Means (best to worst by AP). HDBSCAN recovers the most
coherent per-element instances; this is an oracle-semantics setting and not directly comparable to
Figure~\ref{fig:qual_instance_learned}.}
\label{fig:qual_instance_cluster}
\end{figure*}

Figure~\ref{fig:qual_instance_learned} and Figure~\ref{fig:qual_instance_cluster} show
qualitative instance segmentation results on the four test buildings. Learned methods
(PointGroup, SoftGroup, ISBNet), trained end-to-end from raw point features with no semantic
labels supplied, are shown separately from geometry-based clustering (DBSCAN, Mean Shift,
K-Means, HDBSCAN) evaluated as an oracle given ground-truth semantics. The learned
methods separate instances far less reliably than the oracle-semantic clustering,
fragmenting or missing repetitive facade elements such as windows and blinds; SoftGroup
is the strongest of the three (Table~4). Geometry-based clustering given
ground-truth semantics in Figure~\ref{fig:qual_instance_cluster} instead recovers coherent,
separable instances, with HDBSCAN the most consistent across elements
(Table~5).

\section{Per-Class Instance Segmentation Results}
\label{appendix:instance_perclass}

Table~\ref{tab:instance_perclass} reports AP50 per thing class on the test split for all seven
instance segmentation entries. Learned scores vary widely by class: SoftGroup ranges from 0.001 on arch
to 0.857 on column, ISBNet is strong only on balcony (0.538), and PointGroup peaks on window (0.170).
PointGroup's over-fragmentation (Section~4.2 of the main paper) is especially costly in pooled AP: its 358
vegetation predictions on BLD005 bury correct ones elsewhere, so vegetation AP50 is 0.52 on BLD026 alone but
only 0.034 pooled.

Window is the weakest class for DBSCAN and HDBSCAN despite being the most numerous, because neighbouring panes
merge, whereas detached classes such as balcony and column are recovered far more reliably under HDBSCAN.

\begin{table}[t]
\centering
\caption{\textbf{Per-class AP50 on the test split.} Learned methods receive no semantic labels;
the four clustering methods are given ground-truth semantics and serve as oracle-semantic clustering baselines, so the
two groups are not directly comparable. Classes ordered by instance count. Best clustering result per class in \textbf{bold}.}
\label{tab:instance_perclass}
\resizebox{\columnwidth}{!}{%
\begin{tabular}{lccc|cccc}
\toprule
& \multicolumn{3}{c|}{Learned} & \multicolumn{4}{c}{Clustering (GT semantics)} \\
\cmidrule(lr){2-4}\cmidrule(lr){5-8}
Class & PointGroup & ISBNet & SoftGroup & DBSCAN & HDBSCAN & KMeans & MeanShift \\
\midrule
Window     & 0.170 & 0.009 & 0.220 & 0.002 & \textbf{0.047} & 0.002 & 0.000 \\
Blinds     & 0.003 & 0.005 & 0.162 & 0.100 & \textbf{0.159} & 0.045 & 0.003 \\
Vegetation & 0.034 & 0.152 & 0.245 & \textbf{0.118} & 0.105 & 0.008 & 0.009 \\
Balcony    & 0.000 & 0.538 & 0.264 & 0.528 & \textbf{0.624} & 0.020 & 0.000 \\
Door       & 0.102 & 0.003 & 0.113 & 0.190 & \textbf{0.234} & 0.027 & 0.023 \\
Arch       & 0.004 & 0.000 & 0.001 & 0.218 & \textbf{0.512} & 0.146 & 0.061 \\
Column     & 0.000 & 0.000 & 0.857 & 0.224 & \textbf{0.662} & 0.000 & 0.000 \\
Stairs     & 0.029 & 0.000 & 0.476 & \textbf{0.800} & 0.400 & 0.244 & 0.600 \\
\bottomrule
\end{tabular}%
}
\end{table}

\section{Additional Cross-Dataset Generalization Results}
\label{appendix:crossdomain}

Table~\ref{tab:crossdomain_full} extends the cross-dataset analysis in the main paper to
the complete nine-cell evaluation matrix between HFX3D and ZAHA, the closest published
facade benchmark, where H, Z, and M denote HFX11, ZAHA11, and the two combined (Mixed11).
All experiments use xyz-only input for a feature-neutral comparison,
seed 42, 200 training epochs, and unweighted cross-entropy.

\begin{table*}[!t]
\centering
\caption{\textbf{Cross-dataset mIoU (\%) across all nine train/test combinations.}
H = HFX11, Z = ZAHA11, M = Mixed11, the two datasets combined. The columns cover in-domain
training (H$\to$H, Z$\to$Z), cross-dataset transfer (H$\to$Z, Z$\to$H), transfer to the combined
domain (H$\to$M, Z$\to$M), and mixed-domain training evaluated on each test set (M$\to$H,
M$\to$Z, M$\to$M). All entries use xyz-only input on the shared 11-class taxonomy.
Best per column in \textbf{bold}.}
\label{tab:crossdomain_full}
\resizebox{\textwidth}{!}{%
\begin{tabular}{lccccccccc}
\toprule
\textbf{Model} & \textbf{H$\to$H} & \textbf{Z$\to$Z} & \textbf{H$\to$Z} & \textbf{Z$\to$H} & \textbf{H$\to$M} & \textbf{Z$\to$M} & \textbf{M$\to$H} & \textbf{M$\to$Z} & \textbf{M$\to$M} \\
\midrule
PointNet~\cite{qi2017pointnet}         & 25.73 & 12.83 & 10.69 & 15.73 & 18.18 & 13.95 & 21.02 & 11.92 & 16.82 \\
PointNet++~\cite{qi2017pointnetpp}     & 25.37 & 12.93 &  9.38 & 14.80 & 14.51 & 13.85 & 23.79 & 13.58 & 19.35 \\
DGCNN~\cite{wang2019dynamic}           & 18.00 & 20.17 &  8.15 & 15.11 & 12.80 & 17.39 & 24.86 & 19.97 & 23.16 \\
KPConv~\cite{thomas2019kpconv}         & \textbf{41.59} & \textbf{31.17} & \textbf{12.08} & \textbf{30.15} & \textbf{26.67} & \textbf{30.84} & \textbf{39.76} & \textbf{27.36} & \textbf{32.15} \\
RandLA-Net~\cite{hu2020randla}         &  3.90 &  2.92 &  1.57 &  0.43 &  2.87 &  3.40 &  0.65 &  1.72 &  1.29 \\
Point Transformer v2~\cite{wu2022ptv2} & 16.66 & 14.04 & 11.15 & 18.97 & 13.77 & 17.48 & 17.77 & 12.08 & 14.84 \\
Point Transformer v3~\cite{wu2024ptv3} & 16.44 & 16.32 & 12.31 & 15.69 & 14.15 & 15.99 & 19.44 & 17.09 & 17.91 \\
\bottomrule
\end{tabular}%
}
\end{table*}

\noindent\textbf{Shared label space.}
Table~\ref{tab:label_mapping} gives the mapping from each source taxonomy onto the shared
11-class space. Our vegetation class is dropped because ZAHA has no equivalent, and its points
are removed rather than marked ignore; the remaining eleven of our classes map one-to-one. On
the ZAHA side, molding and interior are dropped for the same reason; interior consists of
indoor points visible through windows and doors. Ground surface and terrain are merged, as they
describe the same physical surface under slightly different annotation conventions, and deco is
merged into other. We also discard ZAHA's colour channel, since it stores one flat colour per
class from the annotation tool rather than measured scanner colour, which is why the comparison
uses xyz-only input.

\begin{table}[t]
\centering
\caption{\textbf{Mapping onto the shared 11-class space} used for all cross-dataset
experiments. Classes marked as dropped have their points removed rather than relabeled
as ignore.}
\label{tab:label_mapping}
\small
\setlength{\tabcolsep}{4pt}
\begin{tabular}{lll}
\toprule
Shared class & HFX3D (12 classes) & ZAHA (15 classes) \\
\midrule
wall        & wall       & wall \\
window      & window     & window \\
door        & door       & door \\
balcony     & balcony    & balcony \\
stairs      & stairs     & stairs \\
roof        & roof       & roof \\
terrain     & terrain    & ground surface + terrain \\
blinds      & blinds     & blinds \\
column      & column     & column \\
arch        & arch       & arch \\
other       & other      & other + deco \\
\midrule
\emph{dropped} & vegetation & molding, interior \\
\bottomrule
\end{tabular}
\end{table}

KPConv achieves the strongest performance across the evaluation matrix.
The cross-dataset experiments nevertheless show a substantial domain gap:
performance drops markedly when models are evaluated outside their training domain.
Transfer is also asymmetric. For example, KPConv trained on ZAHA11 achieves
30.15\% on HFX11, whereas the HFX11-trained model reaches only 12.08\% on ZAHA11.
Mixed-domain training partly mitigates this gap, although its effect varies by architecture.
For instance, DGCNN and Point Transformer v3 improve on HFX11 under mixed training,
whereas KPConv and PointNet remain strongest under their respective in-domain settings.

\begin{figure*}[t]
\centering
\includegraphics[width=\linewidth]{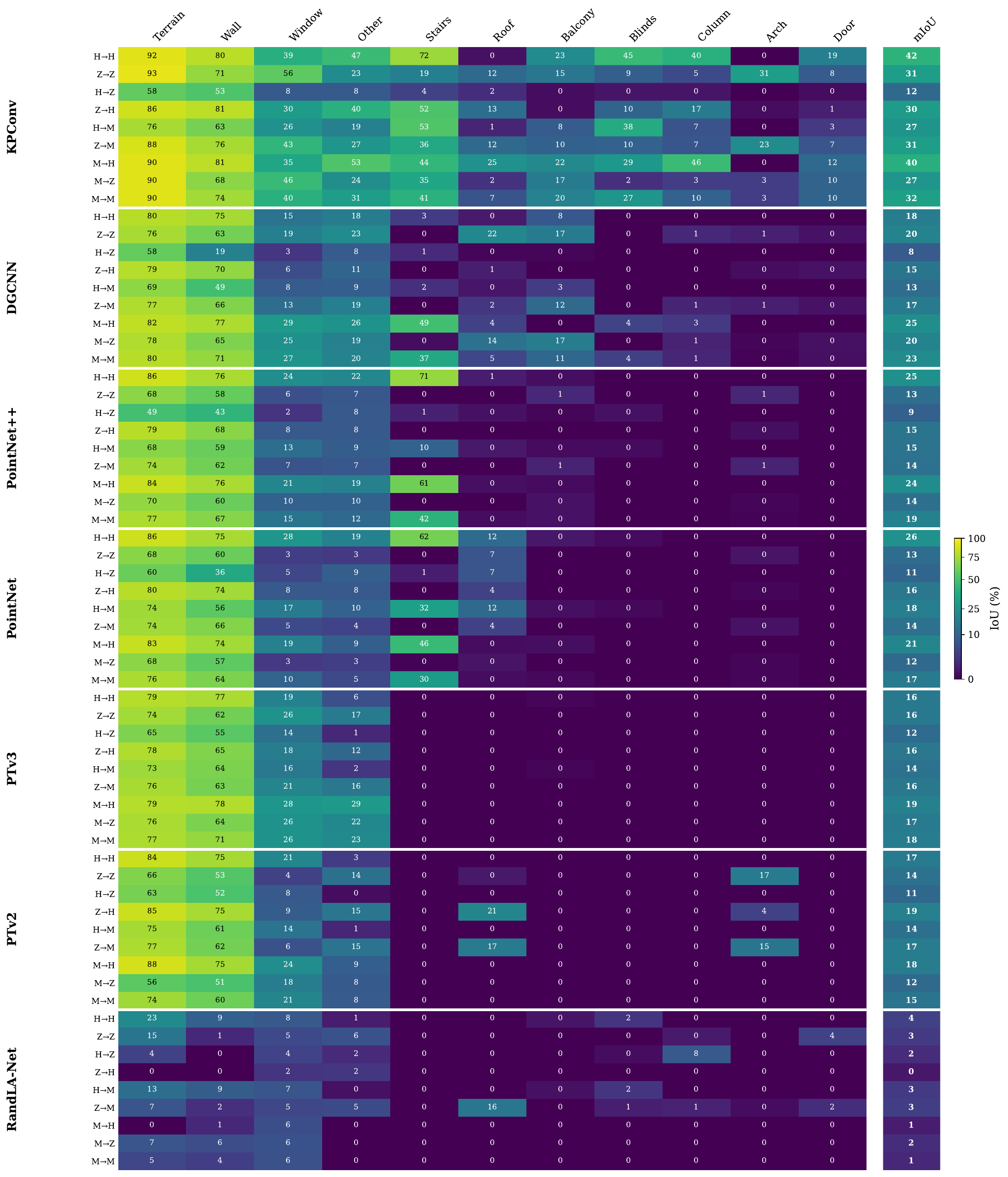}
\caption{\textbf{Per-class IoU (\%) across models, transfer settings, and classes.}
Each model block stacks the nine train$\to$test combinations (H / Z / M = HFX11 / ZAHA11 /
combined); columns are the 11 classes (most- to least-accurate) plus a mIoU column. Cross-dataset
transfer (H$\to$Z, Z$\to$H) trails in-domain.}
\label{fig:crossdomain_perclass}
\end{figure*}

Figure~\ref{fig:crossdomain_perclass} decomposes this matrix into per-class IoU, making the
source of the domain gap explicit: cross-dataset transfer degrades the frequent classes
(wall, terrain, window) only modestly but all but erases the already-fragile rare classes.
Only KPConv retains any signal on classes such as blinds, column, and arch, and RandLA-Net
collapses across nearly every class and transfer setting.

\section{Benchmark Distribution}
\label{sec:distribution}
\begin{figure*}[t]
\centering
\includegraphics[width=\linewidth]{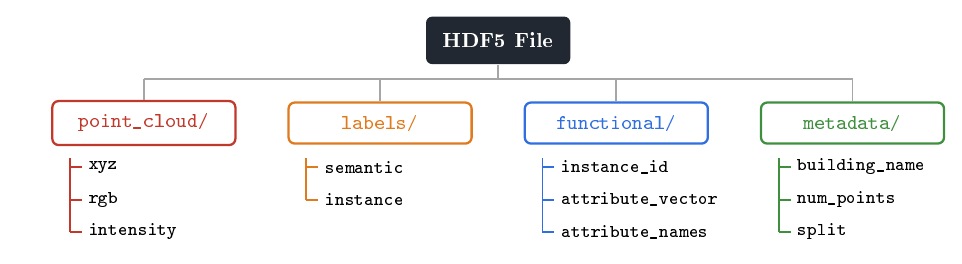}
\caption{\textbf{Representative HDF5 layout.} Each building is stored as one HDF5 file with
four top-level groups (\texttt{point\_cloud}, \texttt{labels}, \texttt{functional},
\texttt{metadata}); \texttt{attribute\_vector} is the final 15-D binary functional vector and
\texttt{attribute\_names} gives its index mapping. The exact field names within each group will
be finalized and documented with the dataset release.}
\label{fig:schema_hierarchy}
\end{figure*}

To encourage adoption and reproducibility, HFX3D
will be released with the following resources:

\noindent\textbf{Dataset.} One HDF5 file per building. Each file is organized into four
top-level groups (Figure~\ref{fig:schema_hierarchy}), covering
raw per-point geometry and colour (\texttt{point\_cloud}), per-point semantic and instance IDs
(\texttt{labels}), per-instance functional attribute fields (\texttt{functional}), and per-file
provenance and split information (\texttt{metadata}). Keeping the functional annotations in the
same file as the points leaves every instance aligned with its geometry without a separate join
step. The four-group structure is fixed; the exact fields within each group will be finalized
and documented with the release.

\begin{figure*}[t]
\centering
\includegraphics[width=\linewidth]{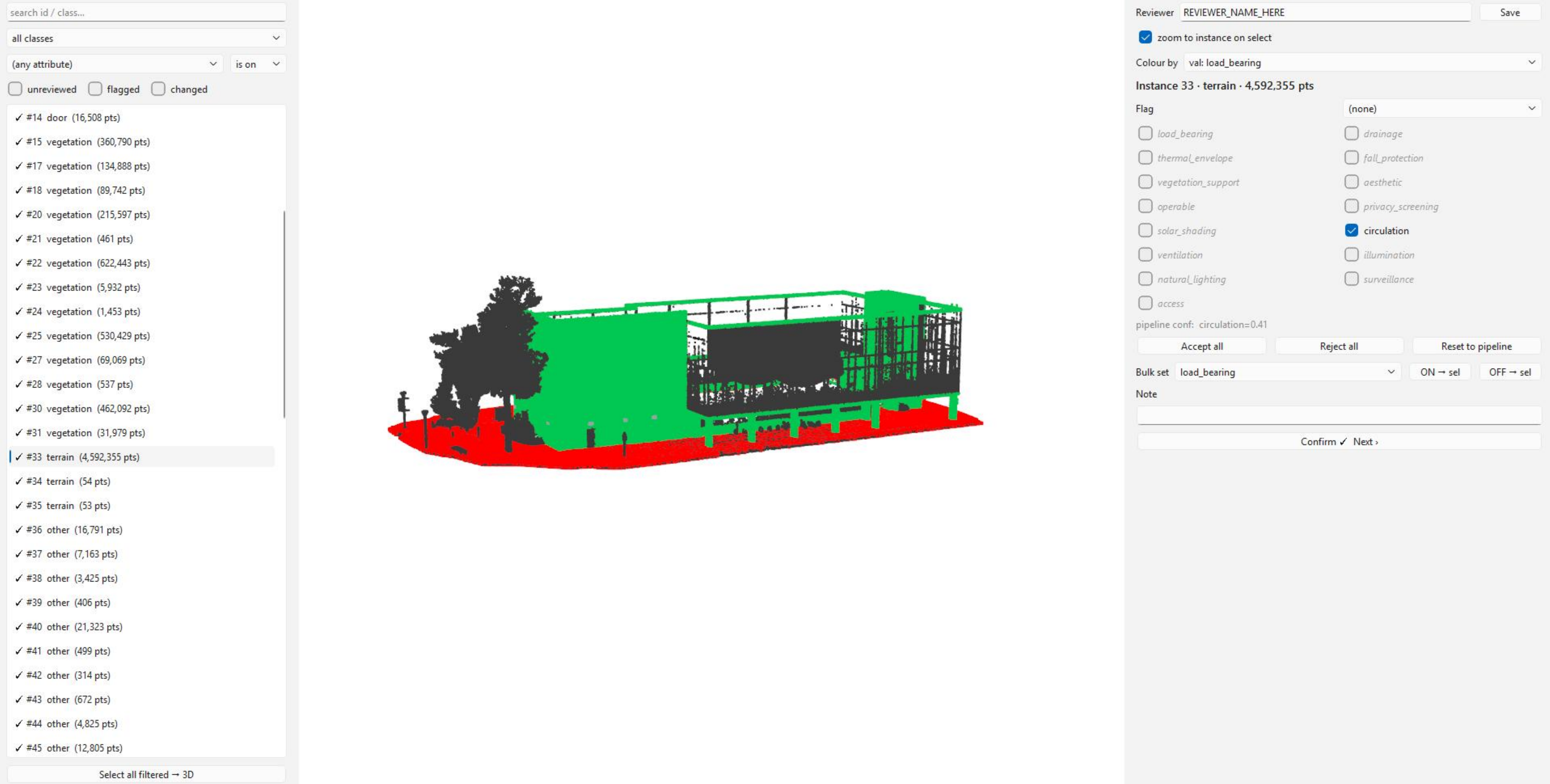}
\caption{\textbf{Functional attribute review interface.} Reviewers select an instance from the
filtered list (left), inspect its geometry in 3D with the current attribute overlay (centre),
and accept, reject, or edit its 15-attribute functional vector against the pipeline-suggested
confidence (right).}
\label{fig:review_tool}
\end{figure*}

\noindent\textbf{Annotation toolkit.} The
ontology-driven rule engine, geometry-guided
refinement scripts, and human review interface
are released as open-source tools to facilitate
extension of functional annotations to new buildings.
Figure~\ref{fig:review_tool} shows the functional attribute review interface used during human
validation (Section~3.4): reviewers select an instance, inspect its pipeline-suggested attribute
confidences, and accept, reject, or bulk-edit the assignment before confirming.

\noindent\textbf{Baseline code.} We will release the adapted training and inference scripts used
for all seven semantic segmentation baselines and three learned instance segmentation baselines,
together with the adapted clustering scripts used for the four geometry-based instance baselines
and the functional attribute labeling scripts, all via the project's GitHub repository
(\url{https://github.com/jijup/hfx3d-benchmark}).

\noindent\textbf{Consent and privacy.} Every building was scanned with the prior consent of its
owner or manager, obtained before acquisition. Scanning was carried out from publicly accessible
ground level around the exterior of each building, and the released data has been reviewed to
exclude interior detail and any personally identifiable content.

\clearpage

   \end{document}